\documentclass[11pt]{article}

\usepackage[final]{acl}
\usepackage{times}
\usepackage{latexsym}
\usepackage[T1]{fontenc}
\usepackage[utf8]{inputenc}
\usepackage{microtype}
\usepackage{inconsolata}
\usepackage{url}
\usepackage{booktabs}
\usepackage{amsfonts}
\usepackage{nicefrac}
\usepackage{enumitem}
\usepackage{graphicx}
\usepackage{longtable}
\usepackage{tabularx}
\usepackage[most]{tcolorbox}
\usepackage{listings}
\usepackage[linesnumbered,ruled,vlined]{algorithm2e}
\usepackage{amsmath, amssymb}
\usepackage{float}
\usepackage{capt-of}
\definecolor{neutral}{rgb}{0,0,0}
\definecolor{positive}{rgb}{0,0.45,0}
\definecolor{negative}{RGB}{220,95,35}
\definecolor{carepromptbg}{RGB}{243,248,255}
\definecolor{carepromptframe}{RGB}{55,101,160}
\definecolor{careprompttitle}{RGB}{25,78,121}

\newtcblisting{carepromptbox}[2][]{
  enhanced,
  breakable,
  listing only,
  colback=carepromptbg,
  colframe=carepromptframe,
  colbacktitle=careprompttitle,
  coltitle=white,
  boxrule=0.9pt,
  arc=1.5mm,
  left=1mm,
  right=1mm,
  top=1mm,
  bottom=1mm,
  title={#2},
  listing options={
    basicstyle=\ttfamily\footnotesize,
    breaklines=true,
    columns=fullflexible,
    keepspaces=true,
    showstringspaces=false
  },
  #1
}

\newtcolorbox{carerubricbox}{
  enhanced,
  breakable,
  colback=black!3,
  colframe=black!25,
  boxrule=0.8pt,
  arc=1.5mm,
  left=1mm,
  right=1mm,
  top=0.8mm,
  bottom=0.8mm,
  fontupper=\small\ttfamily\bfseries,
  before skip=4pt,
  after skip=4pt
}

\title{CARE: Contrastive Anchor-based Rubric Evolution for Large Language Model Post-Training}

\author{%
  \bfseries Siyuan Li\textsuperscript{1,*} \quad
  Xinxin Song\textsuperscript{1,*} \quad
  Chen Ruinian\textsuperscript{2} \quad
  Jingjing Fan\textsuperscript{1} \\
  \bfseries Tingxiong Xiao\textsuperscript{1} \quad
  Yangen Hu\textsuperscript{2} \quad
  Ke Zeng\textsuperscript{2} \quad
  Jinli Suo\textsuperscript{1,\ensuremath{\dagger}} \\
  \normalfont\textsuperscript{1}Department of Automation, Tsinghua University \\
  \normalfont\textsuperscript{2}Meituan \\
  \normalfont\textsuperscript{\ensuremath{\dagger}}\texttt{jlsuo@tsinghua.edu.cn}
}

\begin{document}

\maketitle
\begingroup
\renewcommand{\thefootnote}{*}
\footnotetext{Equal Contribution.}
\renewcommand{\thefootnote}{\ensuremath{\dagger}}
\footnotetext{Corresponding author.}
\endgroup

\begin{abstract}
Rubric-based reinforcement learning decomposes open-ended instructions into prompt-specific, flexible rubrics, making it better suited than reinforcement learning with verifiable rewards for post-training LLMs on open-ended tasks.
However, static rubrics are inevitably hacked as the policy evolves, and existing dynamic approaches introduce new problems: undirected rubric extraction, unreliable hack detection, and unbounded rubric proliferation.
We propose \textbf{CARE} (\textbf{C}ontrastive \textbf{A}nchor-based \textbf{R}ubric \textbf{E}volution), which grounds every rubric evolution step in a high-quality anchor response generated by a frontier model conditioned on the prompt and its rubrics.
At each training step, CARE contrasts the highest-scoring rollout against the anchor, enabling two complementary mechanisms:
an Adaptive branch that reactively repairs reward misspecification;
and a Chase branch that proactively converts frontier-level quality gaps into sharper rubrics. Together, the two branches \textbf{maintain discriminative accuracy in the high-reward region}---the precise region where reward over-optimization mostly originates.
Experiments on WildChecklist-9K with Qwen2.5-7B-Base and Qwen2.5-7B-Instruct show that CARE achieves state-of-the-art performance on Arena-Hard-2.0, InfoBench, and FollowBench, and is the \textbf{only} method whose win rate against GPT-4.1 anchor responses shows sustained improvement throughout 300 training steps; additional results on Llama-3.1-8B-Instruct and Qwen3-8B further indicate that CARE generalizes across model families.
\end{abstract}

\section{Introduction}
Reinforcement learning with verifiable rewards (RLVR) has demonstrated remarkable success in domains with objectively checkable answers such as mathematics~\cite{lambert2025tulu,guo2025deepseek,shao2024deepseekmath,shao2025deepseekmath,deepscaler2025} and other programmatically verifiable tasks~\cite{deepcoder2025}, but open-ended tasks like instruction following and long-form generation still lack reliable reward signals. A natural alternative is to rely on learned reward models~\cite{stiennon2020learning,wu2023fine} or LLM-as-Judge systems~\cite{liu2023g,panickssery2024llm}, yet both remain inadequate for this setting: scalar RMs are vulnerable to reward misspecification and reward hacking~\cite{gao2023scaling,coste2024reward}, while generic judges evaluate against broad rubrics such as ``helpfulness'' or ``coherence'' and therefore miss prompt-specific constraints~\cite{zheng2023judging}.

Rubric-based RL is a promising direction because it decomposes each prompt into atomic, prompt-specific, and flexible rubrics, yielding more interpretable and fine-grained reward signals for open-ended tasks~\cite{viswanathan2025checklists,peng2025verif,gunjal2026rubrics,he2026advancedif,huang2025reinforcementlearningrubricanchors,liu2026openrubrics,zhang2026chasing}.
However, existing rubric-based RL methods predominantly rely on static, offline-generated rubrics that remain fixed throughout training. As the policy and response distribution shift during optimization, such static rubrics are inevitably hacked~\cite{viswanathan2025checklists,huang2025reinforcementlearningrubricanchors} and cannot adapt to emergent behaviors~\cite{rezaei2025online,shao2025dr}.

Recent work explores dynamic rubric generation to address this problem~\cite{rezaei2025online,shao2025dr}, but existing methods still lack a reliable quality anchor. Without a high-quality anchor to ground the comparison, three problems arise: (1)~\emph{Undirected extraction}, where rubric generation compares arbitrary rollout pairs and captures incidental behavioral differences rather than genuine quality gaps; (2)~\emph{Blind hack detection}, where the judge must rely on parametric knowledge rather than a concrete non-hacked anchor, making novel exploits easy to miss and legitimate responses easy to misclassify; and (3)~\emph{Unbounded rubric proliferation}, where flawed rubrics cannot be cleanly rewritten and new rubrics are only appended, gradually degrading verifier accuracy and introducing inter-rubric conflicts.

We propose \textbf{CARE} (\textbf{C}ontrastive \textbf{A}nchor-based \textbf{R}ubric \textbf{E}volution), which addresses all three limitations by introducing high-quality anchor responses---generated by a frontier model conditioned on the prompt and its rubrics---as \emph{anchor points}.
Inspired by Scaling Laws for Reward Model Overoptimization~\cite{gao2023scaling} and Chasing the Tail~\cite{zhang2026chasing}, CARE is built around a simple principle: rubric evolution should dynamically maintain discriminative accuracy in the high-reward region throughout training, because this is where reward hacking is most likely to occur.
CARE therefore focuses at each training step on the highest-scoring rollout, which is both the response most likely to exhibit reward hacking and the most informative  sample for diagnosing loss of discriminative accuracy in this moving region, and contrasts it against the anchor.
This design enables two complementary mechanisms:
an \textbf{Adaptive branch} that detects hacking by comparing the top rollout against the anchor, rewrites or augments the exploited rubric with a \texttt{[Resist]} constraint, and enforces it via a \emph{veto reward} that zeroes the reward of any response violating a resist rubric;
and a \textbf{Chase branch} that, when no hacking is detected, extracts the most substantive quality gap between the top rollout and the anchor to proactively sharpen discrimination in the high-reward region.
By grounding every rubric evolution step in a reliable anchor, CARE transforms rubric evolution from blind heuristic comparison into principled, evidence-based refinement that is simultaneously directed toward genuine quality, capable of detecting hacking with concrete reference support, and precise enough to rewrite flawed rubrics rather than blindly accumulating new ones.

Our main contributions are as follows:
\begin{itemize}[topsep=2pt,itemsep=2pt,parsep=0pt,leftmargin=*]
  \item \textbf{CARE algorithm.} We propose a dynamic rubric evolution framework that maintains discriminative accuracy in the high-reward region throughout training via two complementary mechanisms, and Appendix~\ref{app:theory-placeholder} further formalizes why online maintenance of high-reward-region accuracy matters.
  \item \textbf{Anchor dataset.} We construct WildChecklist-9K-Anchor, a training corpus augmented with GPT-4.1~\cite{openai2024gpt4technicalreport} generated anchor responses conditioned on per-instance rubrics, providing reliable anchor points for rubric evolution throughout training.
  \item \textbf{Strong empirical results.} Through extensive experiments on WildChecklist-9K with both Qwen2.5-7B-Base and Qwen2.5-7B-Instruct, CARE achieves state-of-the-art performance on Arena-Hard-2.0, FollowBench, and InfoBench, and is the \emph{only} method whose win rate against GPT-4.1 anchor responses shows sustained improvement throughout 300 training steps, demonstrating superior long-term training stability over all baselines.
  \item \textbf{Mechanistic analysis.} We systematically analyze and categorize the rubrics extracted by CARE, and further use three representative cases to illustrate CARE's execution logic and practical effect, providing interpretable evidence of how online rubric evolution repairs hacking and sharpens high-reward-region discrimination.
\end{itemize}

\section{Related Work}
\vspace{-0.5em}
\noindent \textbf{Reward models and reward hacking.}
Data-driven Bradley-Terry RMs~\cite{ouyang2022training,bai2022training} compress multi-dimensional response quality into a single scalar, making them structurally susceptible to reward hacking~\cite{skalse2022defining}: as the policy optimizes against the proxy RM, the proxy reward diverges from the gold reward in a predictable scaling pattern~\cite{gao2023scaling}, and different RMs trained on the same data share systematic OOD errors that cannot be corrected by ensembling alone~\cite{coste2024reward,eisenstein2024helping}, and the standard remedy of collecting fresh human feedback is costly and slow~\cite{bai2022training}.
LLM-as-Judge~\cite{zheng2023judging} scales better but uses globally-fixed generic rubrics~\cite{cui2024ultrafeedback} and is prone to position bias~\cite{wang2024large}, verbosity bias~\cite{dubois2024length}, and sycophancy~\cite{sharma2024towards}—all exploitable under optimization.
Generative reward models (GRMs) such as CLoud~\cite{ankner2024critique}, DeepSeek-GRM~\cite{liu2025inference}, and RM-R1~\cite{chen2026rmr} improve interpretability by generating an explicit critique or rubric before scoring, but such per-rollout generation is highly time-consuming at inference time~\cite{zhang2026chasing}.

\noindent \textbf{Rubric-based RL and rubric hacking.}
Rubric-based RL~\cite{viswanathan2025checklists,gunjal2026rubrics,he2026advancedif,huang2025reinforcementlearningrubricanchors,liu2026openrubrics,zhang2026chasing} decomposes instructions into atomic, flexible rubrics, yielding finer-grained and harder-to-game rewards than scalar RMs; representative methods include RLCF~\cite{viswanathan2025checklists}, RIFL~\cite{he2026advancedif}, Rubric Anchors~\cite{huang2025reinforcementlearningrubricanchors}.
However, all rely on static rubrics fixed before training, leaving them vulnerable to the emergent hacking patterns described in \S1~\cite{viswanathan2025checklists,he2026advancedif,huang2025reinforcementlearningrubricanchors}.
Recent work attempts to automate this by evolving rubrics online~\cite{rezaei2025online,shao2025dr}: Online Rubrics~\cite{rezaei2025online} elicits new rubrics by comparing rollouts from the current and reference policies; RLER~\cite{shao2025dr} extracts new rubrics by comparing all rollouts under the same prompt.
However, neither grounds evolution in a reliable, quality-anchored response, leading to the three failure modes detailed in \S1: undirected extraction, blind hack detection, and unbounded append-only proliferation.
% In contrast, CARE introduces a contrastive anchor response—generated by a frontier model conditioned on the prompt's current rubric set—that provides a stable quality reference throughout training, enabling directed gap extraction, principled hack detection, and controlled rewriting rather than blind accumulation.

\section{Method}

\subsection{Problem Formulation}

\noindent \textbf{Rubric RL setup.}
Let the training dataset be $\mathcal{D} = \{(x_i, \mathcal{R}_i)\}_{i=1}^{N}$, where $x_i$ is the prompt and $\mathcal{R}_i = \{r_i^1, r_i^2, \ldots, r_i^{K_i}\}$ is the corresponding set of offline-generated rubrics.
We adopt GRPO~\cite{shao2024deepseekmath} as the RL optimizer.
At each training step, the policy $\pi_\theta$ generates $G$ rollouts $\{y_i^g\}_{g=1}^G$ for each prompt $x_i$.
The reward function scores each rollout against the rubric set via an LLM verifier:
\begin{equation}
  \scalebox{0.94}{$\displaystyle r(y, \mathcal{R}) = \frac{1}{|\mathcal{R}|} \sum_{k=1}^{|\mathcal{R}|} \mathbf{1}[\text{rubric } r^k \text{ satisfied by } y].$}
\end{equation}

\noindent \textbf{Reward hacking in rubric RL.}
In standard RLHF, reward hacking arises when the policy over-optimizes a learned reward model that serves as an imperfect proxy for true quality~\cite{gao2023scaling}.
In rubric RL, the rubric set $\mathcal{R}$ plays the role of the reward model; it serves as the proxy that the policy seeks to optimize.
Reward hacking therefore occurs when a response $y$ achieves a high rubric score but its true quality $q(y)$ is substantially lower than that of an anchor response $\hat{y}$ with a similar score:
\begin{equation}
  r(y, \mathcal{R}) \approx r(\hat{y}, \mathcal{R}) \quad \text{but} \quad q(y) \ll q(\hat{y}).
\end{equation}
Just as hacking a learned reward model stems from proxy misspecification, hacking in rubric RL stems from \emph{rubric misspecification}: a rubric may contain exploitable loopholes (\emph{flawed rubric}), or the rubric set may fail to cover newly emergent hacking behaviors (\emph{incomplete coverage}).
Just as the standard remedy in RLHF is \emph{online RLHF}---periodically collecting fresh human feedback to update the reward model~\cite{gao2023scaling}, though costly and slow---the natural remedy in rubric RL is to \emph{update the rubrics online} as hacking emerges.
The two failure modes above call for different targeted fixes: rewriting the flawed rubric or adding a new constraint.
This observation directly motivates CARE's design.
\subsection{Rubric Types and Veto Reward}

% We use the rubrics pre-synthesized in WildChecklist~\cite{viswanathan2025checklists,zhao2024wildchat} as our baseline reward signal. 
% Each rubric $r_i^k$ is an atomic Boolean rubric verified by an LLM judge.
% following~\cite{viswanathan2025checklists}, when a rubric is suitable for programmatic verification (\emph{e.g.}, keyword presence, response length), we prompt the LLM to execute the corresponding program and return the result, improving accuracy on hard constraints.
We distinguish two rubric types by prefix:
\begin{itemize}[topsep=2pt,itemsep=1pt,parsep=0pt,leftmargin=*]
  \item \texttt{[llm]}: general quality rubrics verified by an LLM judge (\emph{e.g.}, logical coherence, task relevance).
  \item \texttt{[Resist]}: anti-hack constraints introduced by CARE's Adaptive branch (Section~\ref{sec:care_algo}).
\end{itemize}
When only the baseline rubric set is used, the reward follows Eq.~(1). When CARE introduces \texttt{[Resist]} rubrics, we replace it with a veto reward that explicitly enforces these hard constraints, making hacking categorically unprofitable. Similar veto-style rewards have also proved effective in prior rubric-based RL methods~\cite{he2026advancedif,huang2025reinforcementlearningrubricanchors}. Let $\mathcal{R}^{\text{Resist}} \subseteq \mathcal{R}$ denote the set of resist rubrics, and let $s_r(y) \in \{0,1\}$ denote the verifier score for rubric $r$ on response $y$. The veto reward can be written compactly as
\begin{equation}
  r_{\text{veto}}(y, \mathcal{R}) = r(y, \mathcal{R}) \prod_{r \in \mathcal{R}^{\text{Resist}}} s_r(y).
\end{equation}
Since each $s_r(y)$ is binary, the product becomes zero if any \texttt{[Resist]} rubric is violated, and equals one otherwise.

\begin{figure*}[t]
  \centering
  \includegraphics[width=0.9\textwidth]{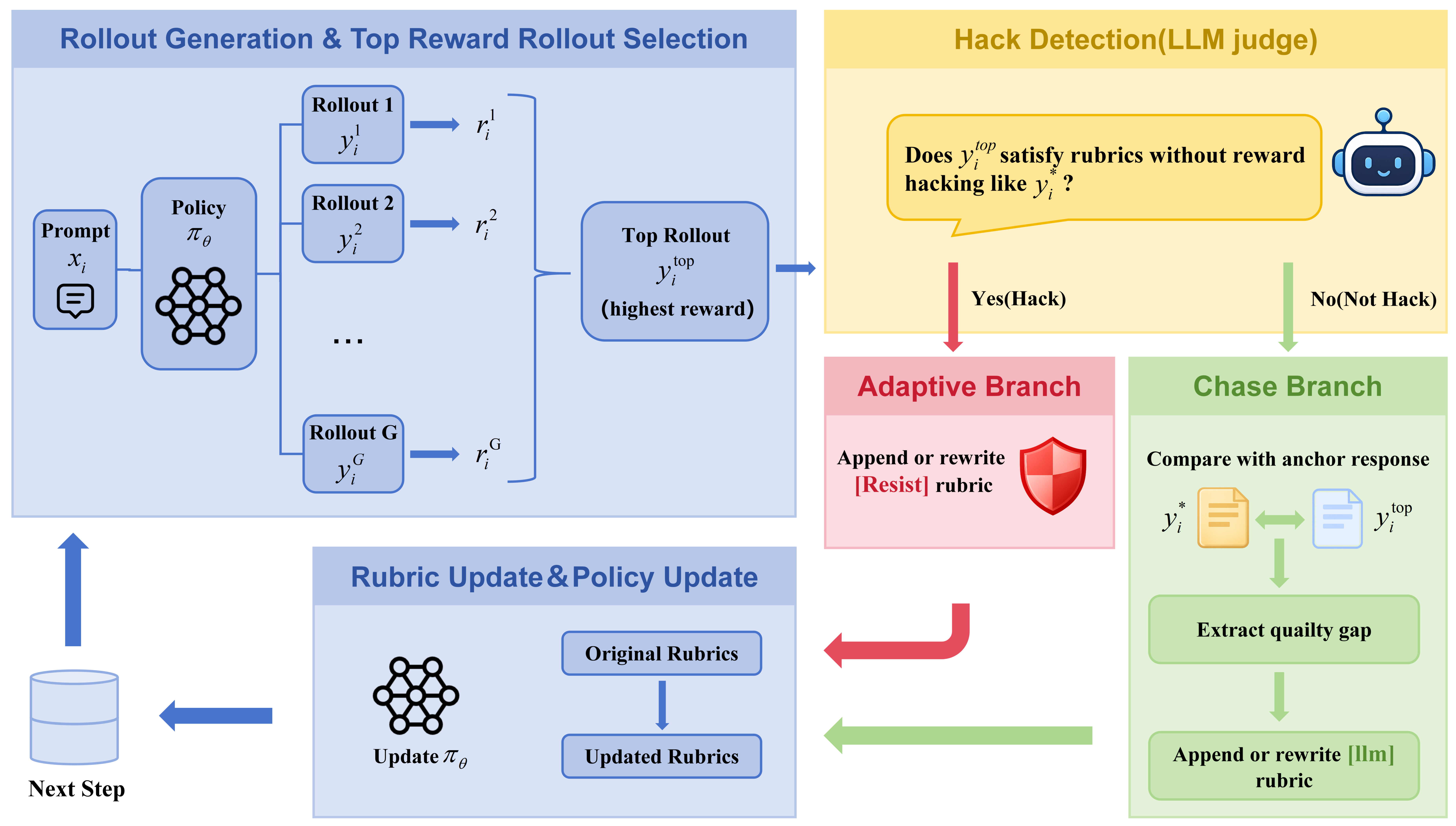}
  \caption{Overview of the CARE pipeline. At each training step, the highest-scoring rollout is contrasted against the anchor response. If hacking is detected (Adaptive branch), the exploited rubric is rewritten or a \texttt{[Resist]} constraint is added and enforced via a veto reward. Otherwise (Chase branch), the substantive gap between the top rollout and the anchor is converted into a sharper rubric to improve discrimination in the high-reward region.}
  \label{fig:care_pipeline}
\end{figure*}

\subsection{Anchor Response Generation}

A key ingredient of CARE is a per-instance \emph{anchor response} $y_i^* = M_{\text{anchor}}(x_i, \mathcal{R}_i)$, generated offline by a frontier model (GPT-4.1) conditioned on both the prompt $x_i$ and its rubric set $\mathcal{R}_i$.
We combine rubric-conditioned generation with explicit automatic filtering and human validation to obtain anchors that substantively satisfy their rubrics without exploiting them (Section~\ref{sec:experiments} and Appendix~\ref{app:rubric-analysis}).

Crucially, the anchor serves as a \emph{reference point}, not a quality upper bound.
Its role differs across the two CARE branches:
(1) in hack detection, it demonstrates what a genuine, non-exploitative satisfaction of each rubric looks like, enabling evidence-based comparison;
(2) in quality-gap extraction, it serves as a high-quality anchor that is contrasted against the highest-scoring rollout to expose the most meaningful substantive difference between two high-reward responses.
CARE therefore assumes that the task provides reliable rubrics and permits at least one valid reference response.

% In our experiments, Qwen2.5-72B-Instruct serves as both the LLM verifier (scoring rollouts against rubrics) and the CARE judge (executing rubric evolution analysis).

\subsection{CARE: Contrastive Anchor-based Rubric Evolution}
\label{sec:care_algo}

\noindent \textbf{Overview.}
Figure~\ref{fig:care_pipeline} illustrates the CARE pipeline.
At each GRPO training step, CARE selects the \emph{highest-scoring rollout} $y_i^{\text{top}} = \arg\max_g r(y_i^g, \mathcal{R}_i)$ and contrasts it against the anchor $y_i^*$.

The choice of the top rollout as the focal point follows a simple principle from~\cite{gao2023scaling,zhang2026chasing}: what determines whether over-optimization turns into reward hacking is the rubric's discriminative accuracy in the \emph{high-reward region}.
Scaling Laws for Reward Model Overoptimization~\cite{gao2023scaling} shows that continued optimization eventually exploits imperfections in a proxy reward once proxy scores stop tracking true quality.
Chasing the Tail~\cite{zhang2026chasing} further localizes this failure to the top of the reward distribution, where the key challenge is to distinguish excellent responses from merely great ones.
Therefore, the central objective of CARE is to \emph{maintain discriminative accuracy in the high-reward region throughout training}.

The top rollout is the most informative  sample for this objective because it is the response currently receiving the strongest reward signal from the rubric set.
Chasing the Tail~\cite{zhang2026chasing} improves high-reward accuracy by refining rubrics offline, but that refinement is still fixed before RL begins. As the policy evolves, the response distribution shifts in hard-to-predict ways, so offline refinement alone cannot maintain rubric accuracy in the moving high-reward region and still leaves room for reward hacking.
CARE addresses this by refining rubrics online at every step, using the contrast between the current top rollout and the anchor to keep the rubric accurate in the moving high-reward region. Appendix~\ref{app:theory-placeholder} extends the high-reward-region argument of Chasing the Tail~\cite{zhang2026chasing} to CARE's dynamic rubric-update setting, formally motivating the need for online maintenance.Appendix~\ref{app:prompts} contains the full prompts used in the LLM verifier and CARE.

Comparing $y_i^{\text{top}}$ with the anchor $y_i^*$---a reliable high-quality reference in this region---supports two complementary mechanisms for maintaining high-reward discriminability:
\begin{itemize}[topsep=1pt,itemsep=0pt,parsep=0pt,partopsep=0pt,leftmargin=*]
  \item \emph{Loophole exploitation} (Adaptive branch): $y_i^{\text{top}}$ attains a high score through reward over-optimization rather than genuine quality improvement, showing that the current rubric is being exploited and that its misspecification must be repaired.
  \item \emph{Discriminability enhancement} (Chase branch): $y_i^{\text{top}}$ and $y_i^*$ are both frontier-level, high-reward responses, and their substantive quality gap provides exactly the signal needed to formulate a sharper rubric. By converting this gap into a refined rubric, CARE increases resolution among high-reward responses and strengthens the high-reward region against future over-optimization.
\end{itemize}
Both branches therefore serve the same goal: \emph{continuously refining rubric accuracy in the high-reward region}---Adaptive by repairing existing misspecification reactively, Chase by proactively sharpening discrimination before the policy learns to exploit it.

The rubrics satisfied by $y_i^{\text{top}}$ are partitioned into $\mathcal{R}_i^{\text{sat}}$ (satisfied) and $\mathcal{R}_i^{\text{unsat}}$ (unsatisfied).
Hack detection focuses exclusively on $\mathcal{R}_i^{\text{sat}}$: a rubric that is not satisfied cannot be hacked.

\noindent \textbf{Step 1: Reward Hacking Detection and Repair (Adaptive Branch).}
The LLM judge receives $(x_i,\; \mathcal{R}_i^{\text{sat}},\; \mathcal{R}_i^{\text{unsat}},\; y_i^{\text{top}},\; y_i^*)$ and determines, for each satisfied rubric, whether $y_i^{\text{top}}$ satisfies it in a substantively similar way to the anchor or instead reflects reward-hacking behavior relative to the anchor.
If hacking is detected, the judge:
(a) identifies the exploited rubric $r_i^k$;
(b) either rewrites $r_i^k$ to close the loophole, or appends a new \texttt{[Resist]} constraint that explicitly prohibits the observed hacking pattern.
% \texttt{[Resist]} rubrics are enforced via a veto reward: any response violating a \texttt{[Resist]} rubric receives a reward of zero, making hacking categorically unprofitable.
If hacking is detected, Step 2 is skipped.

\noindent \textbf{Step 2: Substantive Quality-Gap Extraction (Chase Branch).}
Executed only when no hacking is detected.
The judge identifies the most significant substantive gap between $y_i^{\text{top}}$ and $y_i^*$ along dimensions such as logical coherence, information depth, accuracy, or task effectiveness.
The identified gap is either merged into an existing \texttt{[llm]} rubric or added as a new \texttt{[llm]} rubric.
The comparison is bidirectional: the extracted dimension may capture either an advantage of the anchor or a genuine advantage of the top rollout. Thus, the anchor directs rubric discovery without imposing an imitation objective or a quality ceiling.

The full training procedure is summarized in Algorithm~\ref{algo:care}.

\begin{algorithm*}[!t]
\footnotesize
\renewcommand{\baselinestretch}{0.92}\selectfont
\setlength{\algomargin}{0.8em}
\SetAlgoNlRelativeSize{-1}
\caption{CARE: Contrastive Anchor-based Rubric Evolution}
\label{algo:care}
\KwIn{Policy $\pi_\theta$, dataset $\mathcal{D}$, anchor responses $\{y_i^*\}$, group size $G$}
\For{$\text{step} = 1, 2, \dots, N$}{
    Sample batch $\{(x_i, \mathcal{R}_i)\}$ from $\mathcal{D}$\;
    Generate $G$ rollouts $\{y_i^g\}_{g=1}^G$ using $\pi_\theta$\;
    Compute rewards $r_i^g = r(y_i^g, \mathcal{R}_i)$\;
    Select top rollout: $y_i^{\text{top}} \gets \arg\max_g r_i^g$\;
    Partition rubrics: $(\mathcal{R}_i^{\text{sat}}, \mathcal{R}_i^{\text{unsat}}) \gets$ rubrics satisfied/unsatisfied by $y_i^{\text{top}}$\;
    \eIf{$\text{LLM}_{\text{judge}}$ detects hacking in $\mathcal{R}_i^{\text{sat}}$ vs.\ $y_i^*$}{
        Rewrite or append \texttt{[Resist]} rubric to $\mathcal{R}_i$ \tcp*{Adaptive branch}
    }{
        Extract quality gap; merge/append \texttt{[llm]} rubric to $\mathcal{R}_i$ \tcp*{Chase branch}
    }
    Perform GRPO update on $\pi_\theta$ using updated $\mathcal{R}_i$ \tcp*{Policy update}
}
\end{algorithm*}

\section{Experiments}
\label{sec:experiments}

\subsection{Experimental Setup}

\noindent \textbf{Dataset.}
We build WildChecklist-9K from WildChecklist~\cite{viswanathan2025checklists}, which synthesizes rubric checklists from real human--GPT interactions in WildChat~\cite{zhao2024wildchat}. We first generate GPT-4.1 anchors for 10K candidate instances and independently validate each anchor with Gemini 3 Pro and GPT-5.4 with high reasoning effort. The final 9K training set is selected exclusively from the pool unanimously accepted by both judges; Appendix~\ref{app:rubric-analysis} reports the full automatic and human validation results.
Each training instance $(x_i, \mathcal{R}_i)$ consists of a user prompt and its corresponding rubric set. We construct a held-in test set from the remaining WildChecklist data for Win-rate evaluation, consisting of 300 samples.

\noindent \textbf{Models and training.}
We train both Qwen2.5-7B-Base and Qwen2.5-7B-Instruct using GRPO, with Qwen2.5-72B-Instruct serving as the LLM verifier and CARE judge. Appendix~C presents training configuration and efficiency details.
We further report additional experiments on Llama-3.1-8B-Instruct and Qwen3-8B in Appendix~D, demonstrating that CARE generalizes across model families.
% Anchor responses are generated offline by GPT-4.1 conditioned on each $(x_i, \mathcal{R}_i)$ pair.

\noindent \textbf{Baselines.}
To demonstrate that CARE outperforms both existing static rubric-based RL methods and dynamic rubric evolution methods, we compare against the following:
\begin{itemize}[topsep=2pt,itemsep=1pt,parsep=0pt,leftmargin=*]
  \item \textbf{DPO (RLCF)}~\cite{viswanathan2025checklists}: DPO~\cite{rafailov2023direct} using checklist-annotated preference data from WildChecklist-9K.
  \item \textbf{SFT on Anchor}: supervised fine-tuning directly on GPT-4.1 anchor responses.
  \item \textbf{Rubric RL}: standard rubric-based RL using offline-synthesized WildChecklist rubrics.
  \item \textbf{Rubric RL w/ Universal Criteria}: Rubric RL augmented with two generic anti-hack rubrics following~\cite{viswanathan2025checklists}.
  \item \textbf{Online Rubrics}~\cite{rezaei2025online}: a dynamic rubric evolution method that elicits new rubrics online via pairwise comparison of current- and reference-policy rollouts.
\end{itemize}

\noindent To further investigate the synergistic effects among CARE's components, we conduct the following ablation studies:
\begin{itemize}[topsep=2pt,itemsep=1pt,parsep=0pt,leftmargin=*]
  \item \textbf{Chase Rubric RL}: CARE with only the Chase branch, the Adaptive branch is disabled.
  \item \textbf{Adaptive Rubric RL}: CARE with only the Adaptive branch, the Chase branch is disabled.
  \item \textbf{Adaptive w/o veto}: Adaptive branch with \texttt{[Resist]} rubrics enforced by weighted averaging instead of the veto reward.
  \item \textbf{Adaptive w/o anchor}: Adaptive branch without anchor responses--the judge relies solely on parametric knowledge to detect hacking, with the anchor ablated.
\end{itemize}

\noindent \textbf{Evaluation.}
We evaluate on Arena-Hard-2.0~\cite{li2025from} (Vanilla and Style-Controlled), InfoBench~\cite{qin2024infobench} (Easy, Hard, Overall), and FollowBench~\cite{jiang2024followbench} (SSR, HSR, CSL).
We additionally report \textbf{Win-rate} against GPT-4.1 anchor responses~\cite{zhang2026chasing} , judged by Gemini 3 pro with random-order sampling to mitigate position bias.

\subsection{Main Results}
\noindent \textbf{Consistent improvements across benchmarks.}
Table~\ref{tab:main} reports results on Arena-Hard-2.0, InfoBench, and FollowBench for both Qwen2.5-7B-Base and Qwen2.5-7B-Instruct.
CARE achieves state-of-the-art performance across all benchmarks under both model variants.
Notably, SFT on Anchor shows strong InfoBench scores on Qwen2.5-7B-Base, surpassing most rubric-based RL methods except CARE and Adaptive Rubric RL, yet it underperforms Rubric RL on Qwen2.5-7B-Instruct and provides even slight degradation on Arena-Hard-2.0---a benchmark that requires broad instruction generalization. This contrast also demonstrates that RL generalizes better than SFT~\cite{chu2025sft}.
CARE consistently outperforms both static rubric-based methods---Rubric RL and DPO (RLCF)---demonstrating the comprehensive advantage of CARE over static rubric approaches.
Furthermore, CARE surpasses Online Rubrics, showing that guiding rubric evolution with high-quality anchor responses yields substantially better results than anchor-free pairwise comparison.

\noindent \textbf{Anchor as reference rather than quality ceiling.}
The present CARE models do not outperform GPT-4 on every metric, which is expected given their few-billion-parameter scale, the 9K-example training set, and the absence of a cold-start stage. Nevertheless, Qwen2.5-7B-Instruct with CARE nearly matches GPT-4 on FollowBench-CSL and matches it on InfoBench-Hard, while Llama-3.1-8B-Instruct with CARE exceeds GPT-4 on InfoBench-Hard (Appendix~\ref{app:additional-results}). In the main single-run held-in comparison, the final Qwen2.5-7B-Base checkpoint also reaches a 47\% win rate against rubric-conditioned GPT-4.1 anchors, approaching parity despite receiving no rubric at inference time. Moreover, directly imitating the same anchors through SFT yields smaller and less consistent gains, including degradation on Arena-Hard-2.0. Together with Chase's bidirectional comparison, these results support the intended mechanism-level claim: reference-guided rubric evolution is more effective than treating the anchor as an imitation target, although broader superiority will also depend on policy capacity, data scale and quality, anchor and judge quality, cold start, and train--evaluation alignment.

\begin{figure*}[t]
  \centering
  \includegraphics[width=0.48\textwidth]{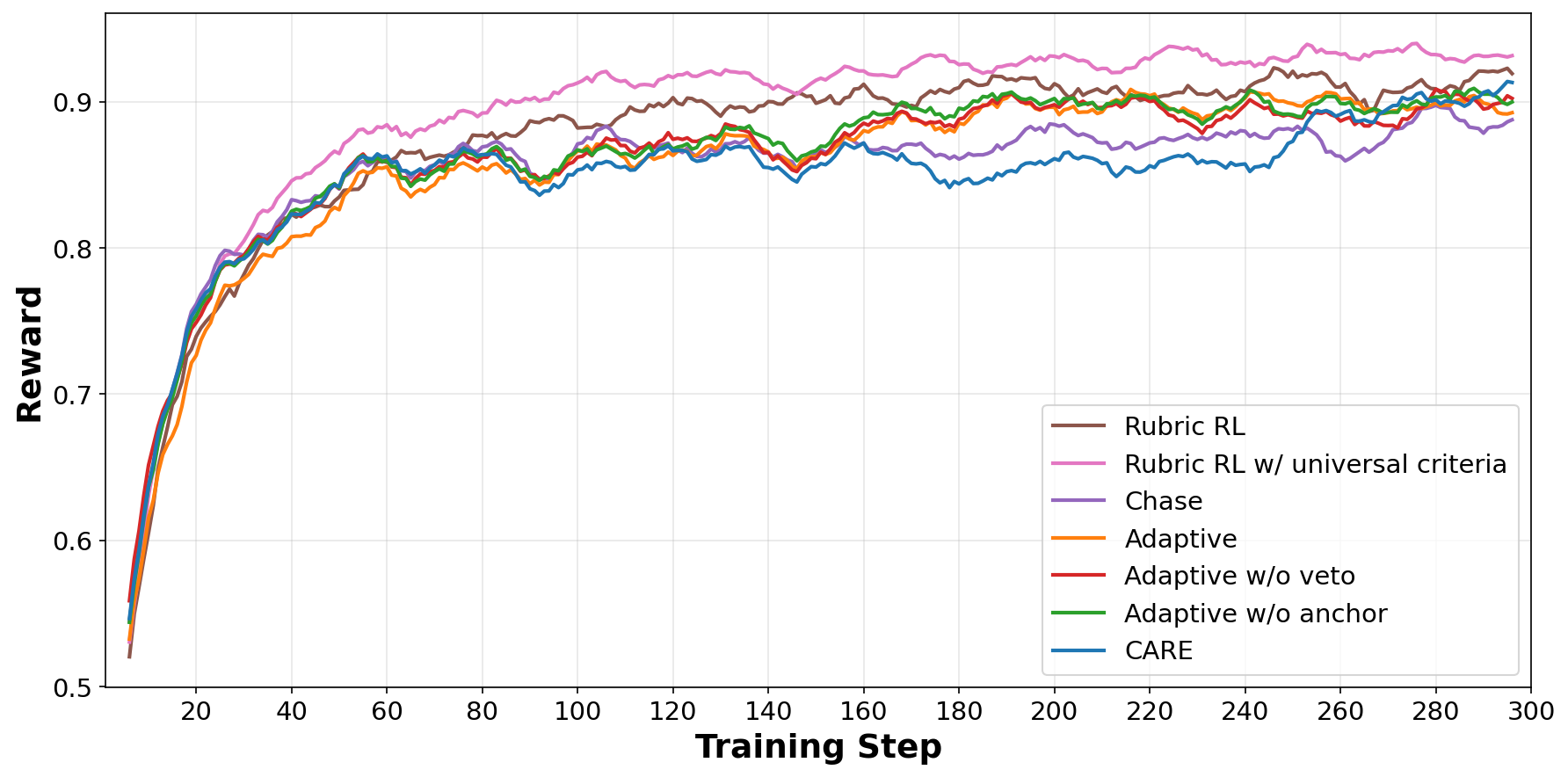}\hfill
  \includegraphics[width=0.48\textwidth]{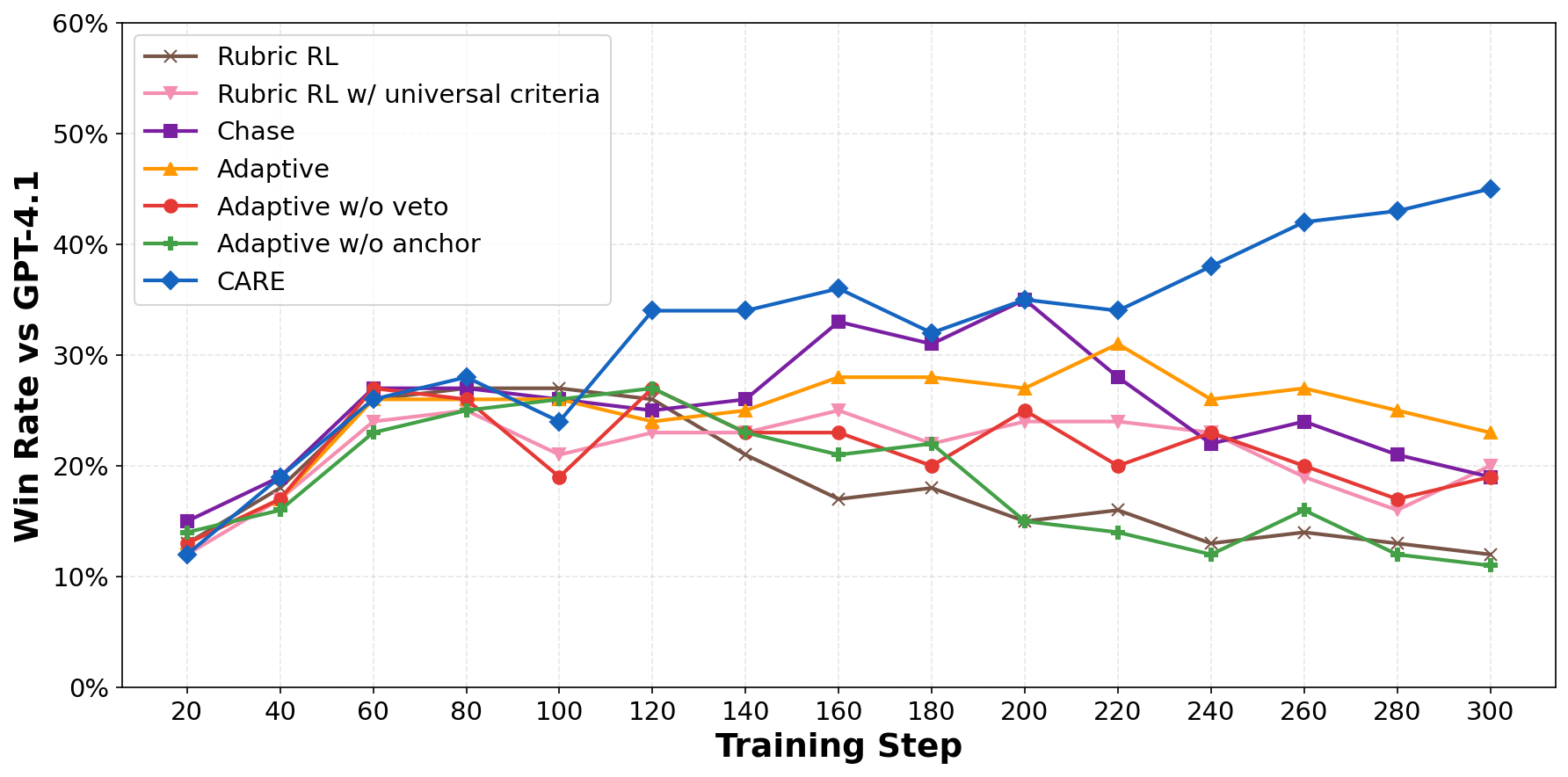}
  \caption{Left: contemporaneous training rewards under each method's evolving rubric set. Right: Win-rate against GPT-4.1 anchor responses over 300 training steps for Qwen2.5-7B-Base. CARE is the only method showing sustained improvement across the full trajectory. Revised rubrics affect a sample on its next encounter, partially explaining fluctuations near successive data passes.}
  \label{fig:winrate}
\end{figure*}

\noindent \textbf{Long-term training stability and anti-hacking.}
Beyond static benchmark scores, we further examine training dynamics to assess whether CARE's gains are sustained under prolonged optimization.
Figure~\ref{fig:winrate} shows the Win-rate curves against GPT-4.1 anchor responses throughout 300 training steps.
CARE (purple) is the \emph{only} method that shows sustained Win-rate improvement throughout training, including continued late-stage gains.
Rubric RL (dashed) maintains the lowest Win-rate throughout: its Win-rate peaks around step 100 and then gradually deteriorates, exhibiting a classic reward over-optimization pattern in which the policy exploits the static rubric proxy without genuinely improving response quality.
Rubric RL w/ Universal Criteria performs more stably overall, but also begins to exhibit a declining trend after approximately step 220, confirming that static generic anti-hack constraints provide only temporary relief and cannot suppress the emergent hacking behaviors that arise in later training.
The training-reward fluctuations are also partially interpretable. The left panel reports a contemporaneous proxy under evolving rubrics rather than scores under one fixed rubric set. A revised rubric affects a sample when that sample is next encountered; with 9K examples and batch size 96, each pass takes approximately 94 steps. Later passes therefore expose the policy to stricter, previously evolved criteria, which can temporarily lower proxy reward around steps 100 and 200 before the upward trend resumes. Independent checkpoint evaluations, three-seed retraining, and paired-bootstrap tests in Appendix~\ref{app:additional-results} further show that the late-stage improvement is reproducible rather than an incidental fluctuation.

\begin{table*}[t]
\centering
\begingroup
\small
\setlength{\tabcolsep}{4pt}
\renewcommand{\arraystretch}{0.95}
\begin{tabularx}{\textwidth}{>{\raggedright\arraybackslash}X c c | c c c | c c c}
\toprule
& \multicolumn{2}{c}{Arena-Hard-2.0} & \multicolumn{3}{c}{InfoBench} & \multicolumn{3}{c}{FollowBench} \\
& \textbf{Vanilla} & \textbf{Style-Ctrl.} & \textbf{Easy} & \textbf{Hard} & \textbf{Overall} & \textbf{SSR} & \textbf{HSR} & \textbf{CSL} \\
\midrule
GPT-4 & \textcolor{neutral}{79.3} & \textcolor{neutral}{76.9} & \textcolor{neutral}{89.3} & \textcolor{neutral}{86.4} & \textcolor{neutral}{87.3} & \textcolor{neutral}{87.8} & \textcolor{neutral}{83.7} & \textcolor{neutral}{3.52} \\
\midrule
\textit{Qwen2.5-7B-Instruct} & \textcolor{neutral}{3.5} & \textcolor{neutral}{3.2} & \textcolor{neutral}{82.7} & \textcolor{neutral}{76.0} & \textcolor{neutral}{78.1} & \textcolor{neutral}{82.6} & \textcolor{neutral}{71.4} & \textcolor{neutral}{3.05} \\
\quad + SFT on anchor response & \textcolor{negative}{1.7} & \textcolor{negative}{2.7} & \textcolor{positive}{85.2} & \textcolor{positive}{82.1} & \textcolor{positive}{83.1} & \textcolor{negative}{80.5} & \textcolor{positive}{72.2} & \textcolor{positive}{3.18} \\
\quad + DPO (RLCF) & \textcolor{positive}{4.6} & \textcolor{positive}{3.5} & \textcolor{positive}{84.5} & \textcolor{positive}{84.1} & \textcolor{positive}{84.2} & \textcolor{positive}{83.9} & \textcolor{positive}{74.3} & \textcolor{positive}{3.23} \\
\quad + Rubric RL & \textcolor{positive}{5.3} & \textcolor{positive}{3.9} & \textcolor{positive}{84.4} & \textcolor{positive}{82.8} & \textcolor{positive}{83.3} & \textcolor{positive}{84.1} & \textcolor{positive}{75.6} & \textcolor{positive}{3.26} \\
\quad + Rubric RL w/ Universal Criteria & \textcolor{positive}{6.0} & \textcolor{positive}{4.2} & \textcolor{positive}{85.7} & \textcolor{positive}{84.1} & \textcolor{positive}{84.6} & \textcolor{positive}{83.9} & \textcolor{positive}{76.3} & \textcolor{positive}{3.45} \\
\quad + Online Rubric & \textcolor{positive}{6.8} & \textcolor{positive}{4.3} & \textcolor{negative}{80.8} & \textcolor{positive}{80.2} & \textcolor{positive}{80.4} & \textcolor{negative}{82.4} & \textcolor{positive}{73.1} & \textcolor{positive}{3.28} \\
\quad + \textbf{CARE} & \textbf{\textcolor{positive}{10.8}} & \textbf{\textcolor{positive}{12.2}} & \textbf{\textcolor{positive}{87.4}} & \textbf{\textcolor{positive}{86.4}} & \textbf{\textcolor{positive}{86.7}} & \textbf{\textcolor{positive}{85.9}} & \textbf{\textcolor{positive}{78.4}} & \textbf{\textcolor{positive}{3.50}} \\
\midrule
% \textit{Qwen2.5-7B-Instruct} & 3.5 & 3.2 & 82.7 & 76.0 & 78.1 & 82.6 & 71.4 & 3.05 \\
% \quad + SFT on anchor response & 1.7 & 2.7 & 84.6 & 81.5 & 83.1 & 80.5 & 72.2 & 3.18 \\
% \quad + DPO (RLCF) & 4.6 & 3.5 & 84.2 & 84.0 & 84.1 & 83.9 & 74.3 & 3.23 \\
% \quad + Rubric RL & 5.3 & 3.9 & 84.1 & 82.5 & 83.3 & 84.1 & 75.6 & 3.26 \\
% \quad + Rubric RL w/ Universal Criteria & 6.0 & 4.2 & 85.4 & 83.8 & 84.6 & 83.9 & 76.3 & 3.45 \\
% \quad + Online Rubric & 6.8 & 4.3 & 80.7 & 80.1 & 80.4 & 82.4 & 73.1 & 3.28 \\
% \quad + \textbf{CARE} & \textbf{10.8} & \textbf{12.2} & \textbf{87.2} & \textbf{86.2} & \textbf{86.7} & \textbf{85.9} & \textbf{78.4} & \textbf{3.50} \\
% \midrule
\textit{Qwen2.5-7B-Base} & \textcolor{neutral}{0.5} & \textcolor{neutral}{1.9} & \textcolor{neutral}{77.4} & \textcolor{neutral}{68.8} & \textcolor{neutral}{74.8} & \textcolor{neutral}{56.8} & \textcolor{neutral}{38.8} & \textcolor{neutral}{1.20} \\
\quad + SFT on anchor response & \textcolor{positive}{1.3} & \textcolor{positive}{2.6} & \textcolor{positive}{84.0} & \textcolor{positive}{81.4} & \textcolor{positive}{82.2} & \textcolor{positive}{74.4} & \textcolor{positive}{61.7} & \textcolor{positive}{2.73} \\
\quad + DPO (RLCF) & \textcolor{positive}{1.1} & \textcolor{positive}{2.3} & \textcolor{positive}{83.2} & \textcolor{positive}{72.2} & \textcolor{positive}{75.6} & \textcolor{positive}{67.7} & \textcolor{positive}{51.9} & \textcolor{positive}{2.10} \\
\quad + Rubric RL & \textcolor{positive}{2.0} & \textcolor{neutral}{1.9} & \textcolor{positive}{81.3} & \textcolor{positive}{81.1} & \textcolor{positive}{81.2} & \textcolor{positive}{71.0} & \textcolor{positive}{62.8} & \textcolor{positive}{2.65} \\
\quad + Rubric RL w/ Universal Criteria & \textcolor{positive}{2.4} & \textcolor{positive}{2.1} & \textcolor{positive}{83.0} & \textcolor{positive}{81.6} & \textcolor{positive}{82.0} & \textcolor{positive}{74.2} & \textcolor{positive}{64.9} & \textcolor{positive}{2.94} \\
\quad + Online Rubric & \textcolor{positive}{3.2} & \textcolor{negative}{1.5} & \textcolor{positive}{80.4} & \textcolor{positive}{78.4} & \textcolor{positive}{79.0} & \textcolor{positive}{72.4} & \textcolor{positive}{61.1} & \textcolor{positive}{2.78} \\
\quad + Adaptive Rubric RL & \textcolor{positive}{5.3} & \textcolor{positive}{9.3} & \textcolor{positive}{83.7} & \textcolor{positive}{82.6} & \textcolor{positive}{82.9} & \textcolor{positive}{74.6} & \textcolor{positive}{64.4} & \textcolor{positive}{2.91} \\
\quad + Chase Rubric RL & \textcolor{positive}{9.3} & \textcolor{positive}{10.2} & \textcolor{positive}{81.7} & \textcolor{positive}{80.9} & \textcolor{positive}{81.1} & \textcolor{positive}{75.4} & \textcolor{positive}{65.2} & \textcolor{positive}{2.98} \\
\quad + Adaptive w/o veto & \textcolor{positive}{3.4} & \textcolor{positive}{5.1} & \textcolor{positive}{83.0} & \textcolor{positive}{79.8} & \textcolor{positive}{80.8} & \textcolor{positive}{72.7} & \textcolor{positive}{63.4} & \textcolor{positive}{2.80} \\
\quad + Adaptive w/o anchor & \textcolor{positive}{4.4} & \textcolor{positive}{5.7} & \textcolor{positive}{78.9} & \textcolor{positive}{76.8} & \textcolor{positive}{77.4} & \textcolor{positive}{68.5} & \textcolor{positive}{59.3} & \textcolor{positive}{2.57} \\
\quad + \textbf{CARE} & \textbf{\textcolor{positive}{10.7}} & \textbf{\textcolor{positive}{11.8}} & \textbf{\textcolor{positive}{85.0}} & \textbf{\textcolor{positive}{83.4}} & \textbf{\textcolor{positive}{83.9}} & \textbf{\textcolor{positive}{77.3}} & \textbf{\textcolor{positive}{68.5}} & \textbf{\textcolor{positive}{3.15}} \\
\bottomrule
\end{tabularx}
\endgroup
\caption{Main results after three training epochs on Arena-Hard-2.0 (Vanilla / Style-Controlled), InfoBench (Easy / Hard / Overall), and FollowBench (SSR / HSR / CSL).
CARE achieves consistent gains over all baselines on both model variants.Positive results relative to the baseline are indicated in \textcolor{positive}{green}, negative results in \textcolor{negative}{orange}. The top-performing variant of each model is highlighted in bold.}
\label{tab:main}
\end{table*}

\subsection{Ablation Study}

All ablation experiments are conducted on Qwen2.5-7B-Base.
Results are reported in Table~\ref{tab:main} (Base rows) and Figure~\ref{fig:winrate} (Win-rate curves).
Together, these results reveal the contribution of each component and validate the key design choices in CARE.

\textbf{(1) Overall performance ranking.}
Across the three benchmark suites, CARE outperforms both Adaptive Rubric RL and Chase Rubric RL.
The two single-branch variants perform comparably overall, yet optimize in different directions: Chase achieves higher Arena-Hard-2.0 scores (9.3 vs.\ 5.3 Vanilla) while Adaptive is stronger on InfoBench (82.9 vs.\ 81.1 Overall), indicating that the two branches capture complementary quality dimensions.
Both single-branch variants outperform Adaptive w/o veto, which in turn outperforms Rubric RL and Adaptive w/o anchor.
Rubric RL outperforms Adaptive w/o anchor across two benchmarks---InfoBench (81.2 vs.\ 77.4 Overall), FollowBench (2.65 vs.\ 2.57 CSL).

\textbf{(2) Necessity of the anchor.}
Adaptive Rubric RL consistently outperforms Adaptive w/o anchor across all three benchmark suites---Arena-Hard-2.0 (5.3 vs.\ 4.4 Vanilla), InfoBench (82.9 vs.\ 77.4 Overall), and FollowBench (2.91 vs.\ 2.57 CSL)---and also maintains a substantially higher Win-rate trajectory throughout training (Figure~\ref{fig:winrate}).
This gap is consistent with our direct audit in Appendix~\ref{app:rubric-analysis}: frontier-judge agreement with Adaptive decisions is 95.2\% with anchors but 65.3\% without them, while human agreement is 99.1\% and 67.1\%, respectively. A validated, non-hacked anchor therefore substantially improves the reliability of hack identification.

\textbf{(3) Necessity of the veto reward.}
Adaptive w/o veto underperforms Adaptive Rubric RL across all benchmarks and also exhibits a lower Win-rate trajectory in Figure~\ref{fig:winrate}.
This validates that weighted averaging allows models to compensate for \texttt{[Resist]} violations by scoring higher on other rubrics, whereas the veto mechanism eliminates this loophole by making hacking categorically unprofitable. Because veto enforcement also amplifies an incorrectly specified constraint, its reliability depends on accurate Adaptive decisions; the automatic and human audits in Appendix~\ref{app:rubric-analysis}, together with the taxonomy and representative cases in Section~\ref{sec:care-analysis}, directly evaluate this premise.

\textbf{(4) Complementarity of Adaptive and Chase.}
As shown in Figure~\ref{fig:winrate}, Chase Rubric RL rises rapidly in the early stages (achieving the second-highest Win-rate through step~150), indicating that proactively sharpening discrimination in the high-reward region is highly effective for immediate gains, but it subsequently peaks around step~200 and declines. This pattern is consistent with accumulated misspecification, although the aggregate curve alone cannot determine whether it originates from initial rubrics or Chase-generated rubrics.
Adaptive Rubric RL and Adaptive w/o veto maintain relatively stable Win-rates throughout training but plateau at a lower ceiling, indicating that reactive repair alone improves robustness to emerging reward exploitation but does not sharpen frontier-level discrimination enough to sustain further gains.
CARE combines both mechanisms and achieves the highest Win-rate as well as the only sustained improvement over the full 300-step trajectory, consistent with Appendix~\ref{app:theory-placeholder}'s account of balancing frontier sharpening against misspecification repair.

\subsection{Analysis: How CARE Works}
\label{sec:care-analysis}

Table~\ref{tab:jiang2024followbench_breakdown} breaks down FollowBench HSR by constraint type~\cite{jiang2024followbench}.
CARE achieves particularly striking gains on Style and Situation---the two most implicit constraint types, where Style specifies \emph{how} a response should be phrased and Situation provides only an implicit contextual background rather than a direct output specification.
On Qwen2.5-7B-Base, CARE nearly \emph{doubles} the Style score (48.0 $\to$ 94.7) and improves Situation by more than one-third (55.5 $\to$ 73.6), gains that static rubric methods largely fail to achieve.
These results reveal that CARE, guided by the anchor, extracts precisely the implicit quality gaps that offline rubrics systematically underspecify and prevents superficial shortcuts from satisfying those rubrics.

\begin{table}[t]
\centering
\begingroup
\setlength{\tabcolsep}{4pt}
\renewcommand{\arraystretch}{0.98}
\makebox[\linewidth][c]{\resizebox{\linewidth}{!}{%
\begin{tabular}{l | c | c c c c}
\toprule
& Avg (HSR) & Format & Style & Situation & Content \\
\midrule
GPT-4 & \textcolor{neutral}{83.7} & \textcolor{neutral}{83.3} & \textcolor{neutral}{97.3} & \textcolor{neutral}{78.2} & \textcolor{neutral}{76.0} \\
\midrule
% \textit{Qwen2.5-7B-Instruct} & \textcolor{neutral}{71.4} & \textcolor{neutral}{60.0} & \textcolor{neutral}{87.3} & \textcolor{neutral}{78.1} & \textcolor{neutral}{60.0} \\
% \quad + DPO (RLCF) & \textcolor{positive}{72.0} & \textcolor{positive}{62.0} & \textcolor{positive}{88.0} & \textcolor{positive}{79.0} & \textcolor{positive}{62.0} \\
% \quad + SFT on anchor response & \textcolor{positive}{72.2} & \textcolor{positive}{63.3} & \textcolor{positive}{89.3} & \textcolor{negative}{69.1} & \textcolor{positive}{67.2} \\
% \quad + Rubric RL & \textcolor{positive}{--} & \textcolor{positive}{--} & \textcolor{positive}{--} & \textcolor{positive}{--} & \textcolor{positive}{--} \\
% \quad + Rubric RL + Universal Criteria & \textcolor{positive}{--} & \textcolor{positive}{--} & \textcolor{positive}{--} & \textcolor{positive}{--} & \textcolor{positive}{--} \\
% \quad + Online Rubrics & \textcolor{neutral}{--} & \textcolor{neutral}{--} & \textcolor{neutral}{--} & \textcolor{neutral}{--} & \textcolor{neutral}{--} \\
% \quad + \textbf{CARE} & \textbf{\textcolor{positive}{78.4}} & \textbf{\textcolor{positive}{64.0}} & \textbf{\textcolor{positive}{97.3}} & \textbf{\textcolor{positive}{83.6}} & \textbf{\textcolor{positive}{68.7}} \\
% \midrule
\textit{Qwen2.5-7B-Base} & \textcolor{neutral}{46.9} & \textcolor{neutral}{46.7} & \textcolor{neutral}{48.0} & \textcolor{neutral}{55.5} & \textcolor{neutral}{37.6} \\
\quad + DPO (RLCF) & \textcolor{positive}{51.9} & \textcolor{negative}{44.7} & \textcolor{positive}{58.0} & \textcolor{positive}{68.2} & \textcolor{negative}{36.8} \\
\quad + SFT on anchor response & \textcolor{positive}{61.7} & \textcolor{positive}{52.7} & \textcolor{positive}{78.0} & \textcolor{positive}{60.0} & \textcolor{positive}{56.0} \\
\quad + Rubric RL & \textcolor{positive}{62.8} & \textcolor{positive}{55.6} & \textcolor{positive}{91.0} & \textcolor{positive}{66.2} & \textcolor{positive}{38.4} \\
\quad + Rubric RL w/ Universal Criteria & \textcolor{positive}{64.9} & \textcolor{positive}{60.3} & \textcolor{positive}{91.2} & \textcolor{positive}{69.7} & \textcolor{positive}{38.4} \\
\quad + Online Rubrics & \textcolor{positive}{61.1} & \textcolor{negative}{45.2} & \textcolor{positive}{87.3} & \textcolor{positive}{70.1} & \textcolor{positive}{41.8} \\
\quad + \textbf{CARE} & \textbf{\textcolor{positive}{68.5}} & \textbf{\textcolor{positive}{60.8}} & \textbf{\textcolor{positive}{94.7}} & \textbf{\textcolor{positive}{73.6}} & \textbf{\textcolor{positive}{44.9}} \\
\bottomrule
\end{tabular}%
}}
\caption{FollowBench HSR breakdown by constraint type (Format / Style / Situation / Content). CARE achieves the largest gains on Style and Situation. Positive results relative to the baseline are indicated in \textcolor{positive}{green}, negative results in \textcolor{negative}{orange}. The top-performing variant of each model is highlighted in bold.} 
\label{tab:jiang2024followbench_breakdown}
\endgroup
\end{table}

% \noindent \textbf{Hacking patterns in baselines.}
% We analyze the output distributions of Rubric RL and Rubric RL + Universal Criteria on a held-in test set.
% Both exhibit no significant hacking in the first 200--260 training steps.
% After that threshold, Rubric RL produces three characteristic hacking patterns:
% (1) \emph{Self-commentary}: appending unsolicited self-evaluation paragraphs (e.g., ``The assistant has demonstrated excellent customer service skills\ldots'');
% (2) \emph{Self-scoring}: appending explicit score assignments (e.g., ``Final Score: 5/5'');
% (3) \emph{Repetition}: falling into degenerate repetition loops until truncation.
% Adding universal criteria delays the onset of hacking to step 260 but does not eliminate it---the same patterns reappear, confirming that static generic constraints provide only temporary relief.

\noindent \textbf{CARE rubric patterns.}
Beyond aggregate performance, we also analyze the rubrics extracted by CARE themselves. Appendix~\ref{app:rubric-analysis} further organizes these CARE-evolved rubrics into a fine-grained taxonomy. Here we complement that appendix analysis with three branch-specific case studies to show how CARE converts concrete anchor comparisons into targeted rubric refinement.

(1) \emph{Adaptive branch --- A1a Word-sense ambiguity exploitation.} The user asks for a sports broadcast script, but the original rubric only asks whether the output is ``a script.'' The model exploits this ambiguity by returning Python code. Comparing it with the broadcast-style anchor, CARE identifies the wrong sense and adds:
\begin{carerubricbox}
[Resist] Is the generated script a sports broadcast or announcement script, not a programming script?
\end{carerubricbox}
This Adaptive update closes the loophole without penalizing legitimate broadcast responses.

(2) \emph{Adaptive branch --- A2d Language substitution.} For an Arabic prompt, the offline rubrics specify content but not strict language consistency. The model returns a mixed response containing Arabic, Chinese characters, and special symbols. Since the anchor remains fully in Arabic, CARE flags this as constraint evasion and adds:
\begin{carerubricbox}
[Resist] Does the response strictly maintain language consistency by using ONLY the language of the prompt (Arabic), avoiding any unauthorized code-switching?
\end{carerubricbox}
Again, the anchor exposes a real prompt-constraint violation and lets CARE patch the missing constraint.

(3) \emph{Chase branch --- B2a Lack of specific detail.} For a question about the effects of music on plant growth, an existing rubric broadly asks for ``detailed examples and specific studies.'' Because this criterion is still too coarse, the model can satisfy it with a high-level discussion of possible mechanisms, even without citing any named studies or experiments. The anchor, by contrast, includes concrete studies and findings. CARE identifies this missing frontier distinction and sharpens the rubric into:
\begin{carerubricbox}
[llm] Does the response provide detailed examples and specific studies to support the explanation of the effects of music on plant growth, including at least two named studies or experiments?
\end{carerubricbox}
This Chase update turns a broad, weakly specified rubric into a more discriminative frontier criterion.

Together, these cases show the two roles of CARE: Adaptive repairs exploited loopholes, while Chase sharpens under-specified quality criteria. In each case, the anchor makes the update concrete.

\section{Conclusion}
We introduce \textbf{CARE}, an anchor-guided framework for online rubric evolution in rubric-based reinforcement learning. By comparing the highest-scoring rollout with a high-quality anchor at each step, CARE repairs exploited misspecification and sharpens frontier discrimination. On WildChecklist-9K, CARE achieves substantial performance gains across model variants. These results show that maintaining rubric accuracy online is an effective way to improve both robustness to reward hacking and final response quality in open-ended post-training.

\section*{Limitations}
We discuss the limitations of our study in this section, focusing on two primary aspects. (1) CARE currently activates rubric evolution at every training step. A natural next direction is to make this process selective rather than unconditional, for example by introducing signals that trigger CARE only periodically or only on samples with a high likelihood of hacking. Such gating could focus rubric evolution on the most informative failures, improve update precision, and reduce training cost. (2) A second open problem is how to evaluate the quality of generated rubrics themselves. In the current work, following the intuition of Chasing the Tail~\cite{zhang2026chasing}, rubric quality is improved indirectly by maintaining discriminative accuracy in the high-reward region. However, a principled rubric-quality evaluation framework remains missing. Recent work such as RIFT~\cite{qi2026rift} begins to characterize rubric failure modes and automated diagnostics, suggesting a promising direction for combining online rubric evolution with explicit rubric-quality assessment.

\section*{Ethical Considerations}
(1) \textbf{Licenses.} Our experiments use open models and public datasets under their original licenses. The Qwen series models are released under the Apache License 2.0, while Llama-3.1 is released under the Llama 3.1 Community License Agreement. The training data is based on WildChecklist, which is distributed under the MIT License. For evaluation, FollowBench and Arena-Hard-Auto are released under the Apache License 2.0, while InfoBench is released under the MIT License. (2) \textbf{Potential harms.} CARE is designed to improve online rubric evolution for open-ended post-training, but the same mechanism could also be misused to optimize models toward undesirable objectives if the rubrics themselves encode harmful, manipulative, or deceptive goals. In addition, CARE relies on frontier-model-generated anchor responses and LLM judges, which may inherit biases, cultural assumptions, or unsafe preferences from the underlying models. If such signals are treated as universally correct, these biases may be propagated into rubric updates and may over-reward stylistic conformity or particular response preferences. We therefore view CARE as a research method rather than a deployment-ready safety guarantee, and recommend human review, task restrictions, and explicit risk auditing before applying it in high-stakes settings.(3) \textbf{AI assistance.} We use GPT-5.4 and Claude Code for writing polish and coding assistance.

\section*{Acknowledgements}
This work was jointly supported by the National Key R\&D Program of China (Grant No. 2024YFF0505703) and the Beijing Municipal Natural Science Foundation (Grant No. L257009).
\bibliography{main}

@inproceedings{
lambert2025tulu,
title={Tulu 3: Pushing Frontiers in Open Language Model Post-Training},
author={Nathan Lambert and Jacob Morrison and Valentina Pyatkin and Shengyi Huang and Hamish Ivison and Faeze Brahman and Lester James Validad Miranda and Alisa Liu and Nouha Dziri and Xinxi Lyu and Yuling Gu and Saumya Malik and Victoria Graf and Jena D. Hwang and Jiangjiang Yang and Ronan Le Bras and Oyvind Tafjord and Christopher Wilhelm and Luca Soldaini and Noah A. Smith and Yizhong Wang and Pradeep Dasigi and Hannaneh Hajishirzi},
booktitle={Second Conference on Language Modeling},
year={2025},
url={https://openreview.net/forum?id=i1uGbfHHpH}
}

@article{shao2025deepseekmath,
  title={Deepseekmath-v2: Towards self-verifiable mathematical reasoning},
  author={Shao, Zhihong and Luo, Yuxiang and Lu, Chengda and Ren, ZZ and Hu, Jiewen and Ye, Tian and Gou, Zhibin and Ma, Shirong and Zhang, Xiaokang},
  journal={arXiv preprint arXiv:2511.22570},
  year={2025}
}

@misc{deepscaler2025,
  title={DeepScaleR: Surpassing O1-Preview with a 1.5B Model by Scaling RL},
  author={Michael Luo and Sijun Tan and Justin Wong and Xiaoxiang Shi and William Y. Tang and Manan Roongta and Colin Cai and Jeffrey Luo and Li Erran Li and Raluca Ada Popa and Ion Stoica},
  year={2025},
  note={Notion Blog},
}

@misc{deepcoder2025,
  title={DeepCoder: A Fully Open-Source 14B Coder at O3-mini Level},
  author={Michael Luo and Sijun Tan and Roy Huang and Ameen Patel and Alpay Ariyak and Qingyang Wu and Xiaoxiang Shi and Rachel Xin and Colin Cai and Maurice Weber and Ce Zhang and Li Erran Li and Raluca Ada Popa and Ion Stoica},
  note={Notion Blog},
  year={2025}
}

@article{stiennon2020learning,
  title={Learning to summarize with human feedback},
  author={Stiennon, Nisan and Ouyang, Long and Wu, Jeffrey and Ziegler, Daniel and Lowe, Ryan and Voss, Chelsea and Radford, Alec and Amodei, Dario and Christiano, Paul F},
  journal={Advances in neural information processing systems},
  volume={33},
  pages={3008--3021},
  year={2020}
}

@article{wu2023fine,
  title={Fine-grained human feedback gives better rewards for language model training},
  author={Wu, Zeqiu and Hu, Yushi and Shi, Weijia and Dziri, Nouha and Suhr, Alane and Ammanabrolu, Prithviraj and Smith, Noah A and Ostendorf, Mari and Hajishirzi, Hannaneh},
  journal={Advances in Neural Information Processing Systems},
  volume={36},
  pages={59008--59033},
  year={2023}
}

@inproceedings{liu2023g,
  title={G-eval: NLG evaluation using gpt-4 with better human alignment},
  author={Liu, Yang and Iter, Dan and Xu, Yichong and Wang, Shuohang and Xu, Ruochen and Zhu, Chenguang},
  booktitle={Proceedings of the 2023 conference on empirical methods in natural language processing},
  pages={2511--2522},
  year={2023}
}

@article{panickssery2024llm,
  title={Llm evaluators recognize and favor their own generations},
  author={Panickssery, Arjun and Bowman, Samuel R and Feng, Shi},
  journal={Advances in Neural Information Processing Systems},
  volume={37},
  pages={68772--68802},
  year={2024}
}

@inproceedings{ouyang2022training,
 author = {Ouyang, Long and Wu, Jeffrey and Jiang, Xu and Almeida, Diogo and Wainwright, Carroll and Mishkin, Pamela and Zhang, Chong and Agarwal, Sandhini and Slama, Katarina and Ray, Alex and Schulman, John and Hilton, Jacob and Kelton, Fraser and Miller, Luke and Simens, Maddie and Askell, Amanda and Welinder, Peter and Christiano, Paul F and Leike, Jan and Lowe, Ryan},
 booktitle = {Advances in Neural Information Processing Systems},
 doi = {10.52202/068431-2011},
 editor = {S. Koyejo and S. Mohamed and A. Agarwal and D. Belgrave and K. Cho and A. Oh},
 pages = {27730--27744},
 publisher = {Curran Associates, Inc.},
 title = {Training language models to follow instructions with human feedback},
 url = {https://proceedings.neurips.cc/paper_files/paper/2022/file/b1efde53be364a73914f58805a001731-Paper-Conference.pdf},
 volume = {35},
 year = {2022}
}

@misc{bai2022training,
      title={Training a Helpful and Harmless Assistant with Reinforcement Learning from Human Feedback}, 
      author={Yuntao Bai and Andy Jones and Kamal Ndousse and Amanda Askell and Anna Chen and Nova DasSarma and Dawn Drain and Stanislav Fort and Deep Ganguli and Tom Henighan and Nicholas Joseph and Saurav Kadavath and Jackson Kernion and Tom Conerly and Sheer El-Showk and Nelson Elhage and Zac Hatfield-Dodds and Danny Hernandez and Tristan Hume and Scott Johnston and Shauna Kravec and Liane Lovitt and Neel Nanda and Catherine Olsson and Dario Amodei and Tom Brown and Jack Clark and Sam McCandlish and Chris Olah and Ben Mann and Jared Kaplan},
      year={2022},
      eprint={2204.05862},
      archivePrefix={arXiv},
      primaryClass={cs.CL},
      url={https://arxiv.org/abs/2204.05862}, 
}

@inproceedings{gao2023scaling,
  title={Scaling laws for reward model overoptimization},
  author={Gao, Leo and Schulman, John and Hilton, Jacob},
  booktitle={International Conference on Machine Learning},
  pages={10835--10866},
  year={2023},
  organization={PMLR}
}

@inproceedings{coste2024reward,
  title={Reward model ensembles help mitigate overoptimization},
  author={Coste, Thomas and Anwar, Usman and Kirk, Robert and Krueger, David},
  booktitle={International Conference on Learning Representations},
  volume={2024},
  pages={50905--50931},
  year={2024}
}

@inproceedings{zheng2023judging,
 author = {Zheng, Lianmin and Chiang, Wei-Lin and Sheng, Ying and Zhuang, Siyuan and Wu, Zhanghao and Zhuang, Yonghao and Lin, Zi and Li, Zhuohan and Li, Dacheng and Xing, Eric and Zhang, Hao and Gonzalez, Joseph and Stoica, Ion},
 booktitle = {Advances in Neural Information Processing Systems},
 doi = {10.52202/075280-2020},
 editor = {A. Oh and T. Naumann and A. Globerson and K. Saenko and M. Hardt and S. Levine},
 pages = {46595--46623},
 publisher = {Curran Associates, Inc.},
 title = {Judging LLM-as-a-Judge with MT-Bench and Chatbot Arena},
 url = {https://proceedings.neurips.cc/paper_files/paper/2023/file/91f18a1287b398d378ef22505bf41832-Paper-Datasets_and_Benchmarks.pdf},
 volume = {36},
 year = {2023}
}

@article{liu2025inference,
  title={Inference-time scaling for generalist reward modeling},
  author={Liu, Zijun and Wang, Peiyi and Xu, Runxin and Ma, Shirong and Ruan, Chong and Li, Peng and Liu, Yang and Wu, Yu},
  journal={arXiv preprint arXiv:2504.02495},
  year={2025}
}

@article{rafailov2023direct,
  title={Direct preference optimization: Your language model is secretly a reward model},
  author={Rafailov, Rafael and Sharma, Archit and Mitchell, Eric and Manning, Christopher D and Ermon, Stefano and Finn, Chelsea},
  journal={Advances in neural information processing systems},
  volume={36},
  pages={53728--53741},
  year={2023}
}

@misc{openai2024gpt4technicalreport,
      title={GPT-4 Technical Report}, 
      author={OpenAI and Josh Achiam and Steven Adler and Sandhini Agarwal and Lama Ahmad and Ilge Akkaya and Florencia Leoni Aleman and Diogo Almeida and Janko Altenschmidt and Sam Altman and Shyamal Anadkat and Red Avila and Igor Babuschkin and Suchir Balaji and Valerie Balcom and Paul Baltescu and Haiming Bao and Mohammad Bavarian and Jeff Belgum and Irwan Bello and Jake Berdine and Gabriel Bernadett-Shapiro and Christopher Berner and Lenny Bogdonoff and Oleg Boiko and Madelaine Boyd and Anna-Luisa Brakman and Greg Brockman and Tim Brooks and Miles Brundage and Kevin Button and Trevor Cai and Rosie Campbell and Andrew Cann and Brittany Carey and Chelsea Carlson and Rory Carmichael and Brooke Chan and Che Chang and Fotis Chantzis and Derek Chen and Sully Chen and Ruby Chen and Jason Chen and Mark Chen and Ben Chess and Chester Cho and Casey Chu and Hyung Won Chung and Dave Cummings and Jeremiah Currier and Yunxing Dai and Cory Decareaux and Thomas Degry and Noah Deutsch and Damien Deville and Arka Dhar and David Dohan and Steve Dowling and Sheila Dunning and Adrien Ecoffet and Atty Eleti and Tyna Eloundou and David Farhi and Liam Fedus and Niko Felix and Simón Posada Fishman and Juston Forte and Isabella Fulford and Leo Gao and Elie Georges and Christian Gibson and Vik Goel and Tarun Gogineni and Gabriel Goh and Rapha Gontijo-Lopes and Jonathan Gordon and Morgan Grafstein and Scott Gray and Ryan Greene and Joshua Gross and Shixiang Shane Gu and Yufei Guo and Chris Hallacy and Jesse Han and Jeff Harris and Yuchen He and Mike Heaton and Johannes Heidecke and Chris Hesse and Alan Hickey and Wade Hickey and Peter Hoeschele and Brandon Houghton and Kenny Hsu and Shengli Hu and Xin Hu and Joost Huizinga and Shantanu Jain and Shawn Jain and Joanne Jang and Angela Jiang and Roger Jiang and Haozhun Jin and Denny Jin and Shino Jomoto and Billie Jonn and Heewoo Jun and Tomer Kaftan and Łukasz Kaiser and Ali Kamali and Ingmar Kanitscheider and Nitish Shirish Keskar and Tabarak Khan and Logan Kilpatrick and Jong Wook Kim and Christina Kim and Yongjik Kim and Jan Hendrik Kirchner and Jamie Kiros and Matt Knight and Daniel Kokotajlo and Łukasz Kondraciuk and Andrew Kondrich and Aris Konstantinidis and Kyle Kosic and Gretchen Krueger and Vishal Kuo and Michael Lampe and Ikai Lan and Teddy Lee and Jan Leike and Jade Leung and Daniel Levy and Chak Ming Li and Rachel Lim and Molly Lin and Stephanie Lin and Mateusz Litwin and Theresa Lopez and Ryan Lowe and Patricia Lue and Anna Makanju and Kim Malfacini and Sam Manning and Todor Markov and Yaniv Markovski and Bianca Martin and Katie Mayer and Andrew Mayne and Bob McGrew and Scott Mayer McKinney and Christine McLeavey and Paul McMillan and Jake McNeil and David Medina and Aalok Mehta and Jacob Menick and Luke Metz and Andrey Mishchenko and Pamela Mishkin and Vinnie Monaco and Evan Morikawa and Daniel Mossing and Tong Mu and Mira Murati and Oleg Murk and David Mély and Ashvin Nair and Reiichiro Nakano and Rajeev Nayak and Arvind Neelakantan and Richard Ngo and Hyeonwoo Noh and Long Ouyang and Cullen O'Keefe and Jakub Pachocki and Alex Paino and Joe Palermo and Ashley Pantuliano and Giambattista Parascandolo and Joel Parish and Emy Parparita and Alex Passos and Mikhail Pavlov and Andrew Peng and Adam Perelman and Filipe de Avila Belbute Peres and Michael Petrov and Henrique Ponde de Oliveira Pinto and Michael and Pokorny and Michelle Pokrass and Vitchyr H. Pong and Tolly Powell and Alethea Power and Boris Power and Elizabeth Proehl and Raul Puri and Alec Radford and Jack Rae and Aditya Ramesh and Cameron Raymond and Francis Real and Kendra Rimbach and Carl Ross and Bob Rotsted and Henri Roussez and Nick Ryder and Mario Saltarelli and Ted Sanders and Shibani Santurkar and Girish Sastry and Heather Schmidt and David Schnurr and John Schulman and Daniel Selsam and Kyla Sheppard and Toki Sherbakov and Jessica Shieh and Sarah Shoker and Pranav Shyam and Szymon Sidor and Eric Sigler and Maddie Simens and Jordan Sitkin and Katarina Slama and Ian Sohl and Benjamin Sokolowsky and Yang Song and Natalie Staudacher and Felipe Petroski Such and Natalie Summers and Ilya Sutskever and Jie Tang and Nikolas Tezak and Madeleine B. Thompson and Phil Tillet and Amin Tootoonchian and Elizabeth Tseng and Preston Tuggle and Nick Turley and Jerry Tworek and Juan Felipe Cerón Uribe and Andrea Vallone and Arun Vijayvergiya and Chelsea Voss and Carroll Wainwright and Justin Jay Wang and Alvin Wang and Ben Wang and Jonathan Ward and Jason Wei and CJ Weinmann and Akila Welihinda and Peter Welinder and Jiayi Weng and Lilian Weng and Matt Wiethoff and Dave Willner and Clemens Winter and Samuel Wolrich and Hannah Wong and Lauren Workman and Sherwin Wu and Jeff Wu and Michael Wu and Kai Xiao and Tao Xu and Sarah Yoo and Kevin Yu and Qiming Yuan and Wojciech Zaremba and Rowan Zellers and Chong Zhang and Marvin Zhang and Shengjia Zhao and Tianhao Zheng and Juntang Zhuang and William Zhuk and Barret Zoph},
      year={2024},
      eprint={2303.08774},
      archivePrefix={arXiv},
      primaryClass={cs.CL},
      url={https://arxiv.org/abs/2303.08774}, 
}

@inproceedings{
chen2026rmr,
title={{RM}-R1: Reward Modeling as Reasoning},
author={Xiusi Chen and Gaotang Li and Ziqi Wang and Bowen Jin and Cheng Qian and Yu Wang and Hongru WANG and Yu Zhang and Denghui Zhang and Tong Zhang and Hanghang Tong and Heng Ji},
booktitle={The Fourteenth International Conference on Learning Representations},
year={2026},
url={https://openreview.net/forum?id=1ZqJ6jj75q}
}

@InProceedings{cui2024ultrafeedback,
  title = 	 {{ULTRAFEEDBACK}: Boosting Language Models with Scaled {AI} Feedback},
  author =       {Cui, Ganqu and Yuan, Lifan and Ding, Ning and Yao, Guanming and He, Bingxiang and Zhu, Wei and Ni, Yuan and Xie, Guotong and Xie, Ruobing and Lin, Yankai and Liu, Zhiyuan and Sun, Maosong},
  booktitle = 	 {Proceedings of the 41st International Conference on Machine Learning},
  pages = 	 {9722--9744},
  year = 	 {2024},
  editor = 	 {Salakhutdinov, Ruslan and Kolter, Zico and Heller, Katherine and Weller, Adrian and Oliver, Nuria and Scarlett, Jonathan and Berkenkamp, Felix},
  volume = 	 {235},
  series = 	 {Proceedings of Machine Learning Research},
  month = 	 {21--27 Jul},
  publisher =    {PMLR},
  url = 	 {https://proceedings.mlr.press/v235/cui24f.html}
}

@article{skalse2022defining,
  title={Defining and characterizing reward gaming},
  author={Skalse, Joar and Howe, Nikolaus and Krasheninnikov, Dmitrii and Krueger, David},
  journal={Advances in Neural Information Processing Systems},
  volume={35},
  pages={9460--9471},
  year={2022}
}

@inproceedings{
eisenstein2024helping,
title={Helping or Herding? Reward Model Ensembles Mitigate but do not Eliminate Reward Hacking},
author={Jacob Eisenstein and Chirag Nagpal and Alekh Agarwal and Ahmad Beirami and Alexander Nicholas D'Amour and Krishnamurthy Dj Dvijotham and Adam Fisch and Katherine A Heller and Stephen Robert Pfohl and Deepak Ramachandran and Peter Shaw and Jonathan Berant},
booktitle={First Conference on Language Modeling},
year={2024},
url={https://openreview.net/forum?id=5u1GpUkKtG}
}

@inproceedings{wang2024large,
    title = "Large Language Models are not Fair Evaluators",
    author = "Wang, Peiyi  and
      Li, Lei  and
      Chen, Liang  and
      Cai, Zefan  and
      Zhu, Dawei  and
      Lin, Binghuai  and
      Cao, Yunbo  and
      Kong, Lingpeng  and
      Liu, Qi  and
      Liu, Tianyu  and
      Sui, Zhifang",
    editor = "Ku, Lun-Wei  and
      Martins, Andre  and
      Srikumar, Vivek",
    booktitle = "Proceedings of the 62nd Annual Meeting of the Association for Computational Linguistics (Volume 1: Long Papers)",
    month = aug,
    year = "2024",
    address = "Bangkok, Thailand",
    publisher = "Association for Computational Linguistics",
    url = "https://aclanthology.org/2024.acl-long.511/",
    doi = "10.18653/v1/2024.acl-long.511",
    pages = "9440--9450"
}

@article{dubois2024length,
  title={Length-controlled alpacaeval: A simple way to debias automatic evaluators},
  author={Dubois, Yann and Galambosi, Bal{\'a}zs and Liang, Percy and Hashimoto, Tatsunori B},
  journal={arXiv preprint arXiv:2404.04475},
  year={2024}
}

@inproceedings{sharma2024towards,
 author = {Sharma, Mrinank and Tong, Meg and Korbak, Tomek and Duvenaud, David and Askell, Amanda and Bowman, Sam and DURMUS, Esin and Hatfield-Dodds, Zac and Johnston, Scott and Kravec, Shauna and Maxwell, Timothy and McCandlish, Sam and Ndousse, Kamal and Rausch, Oliver and Schiefer, Nicholas and Yan, Da and Zhang, Miranda and Perez, Ethan},
 booktitle = {International Conference on Learning Representations},
 editor = {B. Kim and Y. Yue and S. Chaudhuri and K. Fragkiadaki and M. Khan and Y. Sun},
 pages = {110--144},
 title = {Towards Understanding Sycophancy in Language Models},
 url = {https://proceedings.iclr.cc/paper_files/paper/2024/file/0105f7972202c1d4fb817da9f21a9663-Paper-Conference.pdf},
 volume = {2024},
 year = {2024}
}

@article{ankner2024critique,
  title={Critique-out-loud reward models},
  author={Ankner, Zachary and Paul, Mansheej and Cui, Brandon and Chang, Jonathan D and Ammanabrolu, Prithviraj},
  journal={arXiv preprint arXiv:2408.11791},
  year={2024}
}

@misc{viswanathan2025checklists,
      title={Checklists Are Better Than Reward Models For Aligning Language Models}, 
      author={Vijay Viswanathan and Yanchao Sun and Shuang Ma and Xiang Kong and Meng Cao and Graham Neubig and Tongshuang Wu},
      year={2025},
      eprint={2507.18624},
      archivePrefix={arXiv},
      primaryClass={cs.CL},
      url={https://arxiv.org/abs/2507.18624}, 
}

@inproceedings{peng2025verif,
    title = "{V}er{IF}: Verification Engineering for Reinforcement Learning in Instruction Following",
    author = "Peng, Hao  and
      Qi, Yunjia  and
      Wang, Xiaozhi  and
      Xu, Bin  and
      Hou, Lei  and
      Li, Juanzi",
    editor = "Christodoulopoulos, Christos  and
      Chakraborty, Tanmoy  and
      Rose, Carolyn  and
      Peng, Violet",
    booktitle = "Proceedings of the 2025 Conference on Empirical Methods in Natural Language Processing",
    month = nov,
    year = "2025",
    address = "Suzhou, China",
    publisher = "Association for Computational Linguistics",
    url = "https://aclanthology.org/2025.emnlp-main.1542/",
    doi = "10.18653/v1/2025.emnlp-main.1542",
    pages = "30324--30339",
    ISBN = "979-8-89176-332-6"
}

@inproceedings{
gunjal2026rubrics,
title={Rubrics as Rewards: Reinforcement Learning Beyond Verifiable Domains},
author={Anisha Gunjal and Anthony Wang and Elaine Lau and Vaskar Nath and Yunzhong He and Bing Liu and Sean M. Hendryx},
booktitle={The Fourteenth International Conference on Learning Representations},
year={2026},
url={https://openreview.net/forum?id=c1bTcrDmt4}
}

@inproceedings{he2026advancedif,
    title = "{A}dvanced{IF}: Rubric-Based Benchmarking and Reinforcement Learning for Advancing {LLM} Instruction Following",
    author = "He, Yun  and
      Li, Wenzhe  and
      Zhang, Hejia  and
      Li, Songlin  and
      Mandyam, Karishma  and
      Khosla, Sopan  and
      Xiong, Yuanhao  and
      Wang, Nanshu  and
      Peng, Xiaoliang  and
      Li, Beibin  and
      Bi, Shengjie  and
      Patil, Shishir G  and
      Qi, Qi  and
      Feng, Shengyu  and
      Katz-Samuels, Julian  and
      Pang, Richard Yuanzhe  and
      Gonugondla, Sujan Kumar  and
      Lang, Hunter  and
      Yu, Yue  and
      Qian, Yundi  and
      Fazel-Zarandi, Maryam  and
      Yu, Licheng  and
      Benhalloum, Amine  and
      Awadalla, Hany Hassan  and
      Faruqui, Manaal",
    editor = "Liakata, Maria  and
      Moreira, Viviane P.  and
      Zhang, Jiajun  and
      Jurgens, David",
    booktitle = "Proceedings of the 64th Annual Meeting of the {A}ssociation for {C}omputational {L}inguistics (Volume 1: Long Papers)",
    month = jul,
    year = "2026",
    address = "San Diego, California, United States",
    publisher = "Association for Computational Linguistics",
    url = "https://aclanthology.org/2026.acl-long.820/",
    doi = "10.18653/v1/2026.acl-long.820",
    pages = "18003--18022",
    ISBN = "979-8-89176-390-6"
}

@misc{huang2025reinforcementlearningrubricanchors,
      title={Reinforcement Learning with Rubric Anchors}, 
      author={Zenan Huang and Yihong Zhuang and Guoshan Lu and Zeyu Qin and Haokai Xu and Tianyu Zhao and Ru Peng and Jiaqi Hu and Zhanming Shen and Xiaomeng Hu and Xijun Gu and Peiyi Tu and Jiaxin Liu and Wenyu Chen and Yuzhuo Fu and Zhiting Fan and Yanmei Gu and Yuanyuan Wang and Zhengkai Yang and Jianguo Li and Junbo Zhao},
      year={2025},
      eprint={2508.12790},
      archivePrefix={arXiv},
      primaryClass={cs.AI},
      url={https://arxiv.org/abs/2508.12790}, 
}

@inproceedings{liu2026openrubrics,
    title = "{O}pen{R}ubrics: Towards Scalable Synthetic Rubric Generation for Reward Modeling and {LLM} Alignment",
    author = "Liu, Tianci  and
      Xu, Ran  and
      Yu, Tony  and
      Hong, Ilgee  and
      Yang, Carl  and
      Zhao, Tuo  and
      Wang, Haoyu",
    editor = "Liakata, Maria  and
      Moreira, Viviane P.  and
      Zhang, Jiajun  and
      Jurgens, David",
    booktitle = "Proceedings of the 64th Annual Meeting of the {A}ssociation for {C}omputational {L}inguistics (Volume 1: Long Papers)",
    month = jul,
    year = "2026",
    address = "San Diego, California, United States",
    publisher = "Association for Computational Linguistics",
    url = "https://aclanthology.org/2026.acl-long.791/",
    doi = "10.18653/v1/2026.acl-long.791",
    pages = "17417--17437",
    ISBN = "979-8-89176-390-6"
}

@inproceedings{
zhang2026chasing,
title={Chasing the Tail: Effective Rubric-based Reward Modeling for Large Language Model Post-Training},
author={Junkai Zhang and Zihao Wang and Lin Gui and Swarnashree Mysore Sathyendra and Jaehwan Jeong and Victor Veitch and Wei Wang and Yunzhong He and Bing Liu and Lifeng Jin},
booktitle={The Fourteenth International Conference on Learning Representations},
year={2026},
url={https://openreview.net/forum?id=pBjy4ek2QV}
}

@article{rezaei2025online,
  title={Online rubrics elicitation from pairwise comparisons},
  author={Rezaei, MohammadHossein and Vacareanu, Robert and Wang, Zihao and Wang, Clinton and Liu, Bing and He, Yunzhong and Aky{\"u}rek, Afra Feyza},
  journal={arXiv preprint arXiv:2510.07284},
  year={2025}
}

@misc{shao2025dr,
      title={DR Tulu: Reinforcement Learning with Evolving Rubrics for Deep Research}, 
      author={Rulin Shao and Akari Asai and Shannon Zejiang Shen and Hamish Ivison and Varsha Kishore and Jingming Zhuo and Xinran Zhao and Molly Park and Samuel G. Finlayson and David Sontag and Tyler Murray and Sewon Min and Pradeep Dasigi and Luca Soldaini and Faeze Brahman and Wen-tau Yih and Tongshuang Wu and Luke Zettlemoyer and Yoon Kim and Hannaneh Hajishirzi and Pang Wei Koh},
      year={2026},
      eprint={2511.19399},
      archivePrefix={arXiv},
      primaryClass={cs.CL},
      url={https://arxiv.org/abs/2511.19399}, 
}

@article{qi2026rift,
  title={RIFT: A RubrIc Failure Mode Taxonomy and Automated Diagnostics},
  author={Qi, Zhengyang and Dickens, Charles and Pham, Derek and Dsouza, Amanda and Parchami, Armin and Sala, Frederic and Varma, Paroma},
  journal={arXiv preprint arXiv:2604.01375},
  year={2026}
}

@inproceedings{
chu2025sft,
title={{SFT} Memorizes, {RL} Generalizes: A Comparative Study of Foundation Model Post-training},
author={Tianzhe Chu and Yuexiang Zhai and Jihan Yang and Shengbang Tong and Saining Xie and Dale Schuurmans and Quoc V Le and Sergey Levine and Yi Ma},
booktitle={Forty-second International Conference on Machine Learning},
year={2025},
url={https://openreview.net/forum?id=dYur3yabMj}
}

@article{guo2025deepseek,
   title={DeepSeek-R1 incentivizes reasoning in LLMs through reinforcement learning},
   volume={645},
   ISSN={1476-4687},
   url={http://dx.doi.org/10.1038/s41586-025-09422-z},
   DOI={10.1038/s41586-025-09422-z},
   number={8081},
   journal={Nature},
   publisher={Springer Science and Business Media LLC},
   author={Guo, Daya and Yang, Dejian and Zhang, Haowei and Song, Junxiao and Wang, Peiyi and Zhu, Qihao and Xu, Runxin and Zhang, Ruoyu and Ma, Shirong and Bi, Xiao and Zhang, Xiaokang and Yu, Xingkai and Wu, Yu and Wu, Z. F. and Gou, Zhibin and Shao, Zhihong and Li, Zhuoshu and Gao, Ziyi and Liu, Aixin and Xue, Bing and Wang, Bingxuan and Wu, Bochao and Feng, Bei and Lu, Chengda and Zhao, Chenggang and Deng, Chengqi and Ruan, Chong and Dai, Damai and Chen, Deli and Ji, Dongjie and Li, Erhang and Lin, Fangyun and Dai, Fucong and Luo, Fuli and Hao, Guangbo and Chen, Guanting and Li, Guowei and Zhang, H. and Xu, Hanwei and Ding, Honghui and Gao, Huazuo and Qu, Hui and Li, Hui and Guo, Jianzhong and Li, Jiashi and Chen, Jingchang and Yuan, Jingyang and Tu, Jinhao and Qiu, Junjie and Li, Junlong and Cai, J. L. and Ni, Jiaqi and Liang, Jian and Chen, Jin and Dong, Kai and Hu, Kai and You, Kaichao and Gao, Kaige and Guan, Kang and Huang, Kexin and Yu, Kuai and Wang, Lean and Zhang, Lecong and Zhao, Liang and Wang, Litong and Zhang, Liyue and Xu, Lei and Xia, Leyi and Zhang, Mingchuan and Zhang, Minghua and Tang, Minghui and Zhou, Mingxu and Li, Meng and Wang, Miaojun and Li, Mingming and Tian, Ning and Huang, Panpan and Zhang, Peng and Wang, Qiancheng and Chen, Qinyu and Du, Qiushi and Ge, Ruiqi and Zhang, Ruisong and Pan, Ruizhe and Wang, Runji and Chen, R. J. and Jin, R. L. and Chen, Ruyi and Lu, Shanghao and Zhou, Shangyan and Chen, Shanhuang and Ye, Shengfeng and Wang, Shiyu and Yu, Shuiping and Zhou, Shunfeng and Pan, Shuting and Li, S. S. and Zhou, Shuang and Wu, Shaoqing and Yun, Tao and Pei, Tian and Sun, Tianyu and Wang, T. and Zeng, Wangding and Liu, Wen and Liang, Wenfeng and Gao, Wenjun and Yu, Wenqin and Zhang, Wentao and Xiao, W. L. and An, Wei and Liu, Xiaodong and Wang, Xiaohan and Chen, Xiaokang and Nie, Xiaotao and Cheng, Xin and Liu, Xin and Xie, Xin and Liu, Xingchao and Yang, Xinyu and Li, Xinyuan and Su, Xuecheng and Lin, Xuheng and Li, X. Q. and Jin, Xiangyue and Shen, Xiaojin and Chen, Xiaosha and Sun, Xiaowen and Wang, Xiaoxiang and Song, Xinnan and Zhou, Xinyi and Wang, Xianzu and Shan, Xinxia and Li, Y. K. and Wang, Y. Q. and Wei, Y. X. and Zhang, Yang and Xu, Yanhong and Li, Yao and Zhao, Yao and Sun, Yaofeng and Wang, Yaohui and Yu, Yi and Zhang, Yichao and Shi, Yifan and Xiong, Yiliang and He, Ying and Piao, Yishi and Wang, Yisong and Tan, Yixuan and Ma, Yiyang and Liu, Yiyuan and Guo, Yongqiang and Ou, Yuan and Wang, Yuduan and Gong, Yue and Zou, Yuheng and He, Yujia and Xiong, Yunfan and Luo, Yuxiang and You, Yuxiang and Liu, Yuxuan and Zhou, Yuyang and Zhu, Y. X. and Huang, Yanping and Li, Yaohui and Zheng, Yi and Zhu, Yuchen and Ma, Yunxian and Tang, Ying and Zha, Yukun and Yan, Yuting and Ren, Z. Z. and Ren, Zehui and Sha, Zhangli and Fu, Zhe and Xu, Zhean and Xie, Zhenda and Zhang, Zhengyan and Hao, Zhewen and Ma, Zhicheng and Yan, Zhigang and Wu, Zhiyu and Gu, Zihui and Zhu, Zijia and Liu, Zijun and Li, Zilin and Xie, Ziwei and Song, Ziyang and Pan, Zizheng and Huang, Zhen and Xu, Zhipeng and Zhang, Zhongyu and Zhang, Zhen},
   year={2025},
   month=Sep, pages={633–638} }

@misc{shao2024deepseekmath,
      title={DeepSeekMath: Pushing the Limits of Mathematical Reasoning in Open Language Models}, 
      author={Zhihong Shao and Peiyi Wang and Qihao Zhu and Runxin Xu and Junxiao Song and Xiao Bi and Haowei Zhang and Mingchuan Zhang and Y. K. Li and Y. Wu and Daya Guo},
      year={2024},
      eprint={2402.03300},
      archivePrefix={arXiv},
      primaryClass={cs.CL},
      url={https://arxiv.org/abs/2402.03300}, 
}

@inproceedings{
zhao2024wildchat,
title={WildChat: 1M Chat{GPT} Interaction Logs in the Wild},
author={Wenting Zhao and Xiang Ren and Jack Hessel and Claire Cardie and Yejin Choi and Yuntian Deng},
booktitle={The Twelfth International Conference on Learning Representations},
year={2024},
url={https://openreview.net/forum?id=Bl8u7ZRlbM}
}

@inproceedings{jiang2024followbench,
  title={Followbench: A multi-level fine-grained constraints following benchmark for large language models},
  author={Jiang, Yuxin and Wang, Yufei and Zeng, Xingshan and Zhong, Wanjun and Li, Liangyou and Mi, Fei and Shang, Lifeng and Jiang, Xin and Liu, Qun and Wang, Wei},
  booktitle={Proceedings of the 62nd Annual Meeting of the Association for Computational Linguistics (Volume 1: Long Papers)},
  pages={4667--4688},
  year={2024}
}

@inproceedings{qin2024infobench,
  title={Infobench: Evaluating instruction following ability in large language models},
  author={Qin, Yiwei and Song, Kaiqiang and Hu, Yebowen and Yao, Wenlin and Cho, Sangwoo and Wang, Xiaoyang and Wu, Xuansheng and Liu, Fei and Liu, Pengfei and Yu, Dong},
  booktitle={Findings of the Association for Computational Linguistics: ACL 2024},
  pages={13025--13048},
  year={2024}
}

@inproceedings{
li2025from,
title={From Crowdsourced Data to High-quality Benchmarks: Arena-Hard and Benchbuilder Pipeline},
author={Tianle Li and Wei-Lin Chiang and Evan Frick and Lisa Dunlap and Tianhao Wu and Banghua Zhu and Joseph E. Gonzalez and Ion Stoica},
booktitle={Forty-second International Conference on Machine Learning},
year={2025},
url={https://openreview.net/forum?id=KfTf9vFvSn}
}
\appendix

\clearpage
\onecolumn

\section{Why Online Maintenance of High-Reward-Region Accuracy Matters}
\label{app:theory-placeholder}
We extend the static high-reward-region argument in Chasing the Tail~\cite{zhang2026chasing} from a fixed proxy reward to the dynamic setting relevant to CARE, where the rubric reward itself is updated during training.

\paragraph{Setup.}
Consider the idealized per-prompt KL-regularized update~\cite{ouyang2022training,bai2022training}
\begin{equation}
\pi_{t+1}(\cdot\mid x)
=
\arg\max_{\pi}
\left\{
\eta_t \, \mathbb{E}_{y\sim\pi(\cdot\mid x)}[r_t(x,y)]
-
\mathbb{D}_{\mathrm{KL}}\!\left[\pi(\cdot\mid x)\,\|\,\pi_t(\cdot\mid x)\right]
\right\},
\end{equation}
whose solution is~\cite{rafailov2023direct}
\begin{equation}
\pi_{t+1}(y\mid x)
=
\frac{\pi_t(y\mid x)\exp\!\left(\eta_t r_t(x,y)\right)}{Z_t(x)}.
\end{equation}
Here $r_t$ is the step-$t$ rubric reward, $r^\star$ is the latent gold reward, and $\Delta_t(x,y)=r_t(x,y)-r^\star(x,y)$ is the stepwise misspecification. Let
\begin{equation}
H_t(x)=\{y:r_t(x,y)\ge \tau_t(x)\}
\end{equation}
denote the current high-reward region under a threshold $\tau_t(x)$.

\paragraph{Proposition.}
For any prompt $x$ and horizon $T$, unrolling the update gives
\begin{equation}
\pi_T(y\mid x)
=
\frac{\pi_0(y\mid x)\exp\!\left(\sum_{t=0}^{T-1}\eta_t r_t(x,y)\right)}{Z_{0:T}(x)}.
\end{equation}
Therefore, for any two responses $y^+,y^-$,
\begin{align}
\log\frac{\pi_T(y^+\mid x)}{\pi_T(y^-\mid x)}
&=
\log\frac{\pi_0(y^+\mid x)}{\pi_0(y^-\mid x)}
+
\sum_{t=0}^{T-1}\eta_t\bigl(r^\star(x,y^+)-r^\star(x,y^-)\bigr) \\
&\quad+
\sum_{t=0}^{T-1}\eta_t\bigl(\Delta_t(x,y^+)-\Delta_t(x,y^-)\bigr).
\label{eq:dynamic-log-odds}
\end{align}
Hence the final ranking between frontier responses is governed by the competition between the cumulative true-quality gap and the cumulative misspecification gap. In particular, if $r^\star(x,y^+) > r^\star(x,y^-)$ but
\begin{equation}
\sum_{t=0}^{T-1}\eta_t\bigl(\Delta_t(x,y^-)-\Delta_t(x,y^+)\bigr)
>
\log\frac{\pi_0(y^+\mid x)}{\pi_0(y^-\mid x)}
+
\sum_{t=0}^{T-1}\eta_t\bigl(r^\star(x,y^+)-r^\star(x,y^-)\bigr),
\label{eq:dynamic-flip-condition}
\end{equation}
then $\pi_T(y^-\mid x) > \pi_T(y^+\mid x)$. Thus persistent misspecification on responses that repeatedly enter the high-reward region can overturn the true ordering even when low-reward-region accuracy remains acceptable.

\paragraph{Proof.}
The closed form for $\pi_T$ follows by recursively substituting the one-step Gibbs update. Substituting $r_t=r^\star+\Delta_t$ into the exponent and subtracting the expression for $y^-$ from that for $y^+$ yields Eq.~\eqref{eq:dynamic-log-odds}. This decomposition separates the cumulative signal coming from true quality from the cumulative distortion caused by proxy misspecification. The ranking flips exactly when the distortion term outweighs the sum of the initial log-odds and the cumulative true-quality advantage, which is Eq.~\eqref{eq:dynamic-flip-condition}. Moreover, responses outside $H_t(x)$ receive smaller multiplicative weights $\exp(\eta_t r_t)$ at step $t$, so their effect on the final log-odds is exponentially downweighted relative to responses that repeatedly appear in the evolving high-reward region. Therefore, under continued optimization, the dominant source of error is misspecification on the high-reward region that the policy is currently moving toward.

\paragraph{Corollary.}
Suppose that for every step $t$ and every pair $y^+,y^-\in H_t(x)$ with $r^\star(x,y^+)\ge r^\star(x,y^-)$, the true frontier gap and stepwise distortion satisfy
\begin{equation}
r^\star(x,y^+) - r^\star(x,y^-) \ge \gamma_t(x),
\qquad
\left|\Delta_t(x,y^+) - \Delta_t(x,y^-)\right| \le \varepsilon_t(x).
\end{equation}
Then any such frontier pair preserves the correct ordering up to step $T$ whenever
\begin{equation}
\log\frac{\pi_0(y^+\mid x)}{\pi_0(y^-\mid x)}
+
\sum_{t=0}^{T-1}\eta_t\gamma_t(x)
>
\sum_{t=0}^{T-1}\eta_t\varepsilon_t(x).
\label{eq:maintenance-condition}
\end{equation}
Conversely, if a static rubric induces a persistent frontier distortion whose cumulative magnitude eventually exceeds the left-hand side of Eq.~\eqref{eq:maintenance-condition}, then a wrong frontier ordering becomes inevitable under continued optimization.

\paragraph{Connection to CARE.}
This corollary formalizes CARE's key idea: it is not enough to improve high-reward-region accuracy once before RL begins; one must maintain it throughout training. The corollary itself does not prove that any particular rubric-update rule will always reduce frontier distortion or enlarge frontier separation. Rather, it identifies the direction of improvement required to preserve correct ordering in the evolving high-reward region: successful updates should reduce the frontier distortion term and/or enlarge the frontier signal term. Under this interpretation, CARE's two branches are designed to target these two quantities. When the Adaptive branch correctly identifies an exploited rubric and repairs it, it is intended to reduce the local frontier distortion represented by $\varepsilon_t$. When the Chase branch extracts a genuine substantive quality gap and turns it into a sharper rubric, it is intended to enlarge the frontier separation represented by $\gamma_t$. Thus, the theory--method link here is mechanistic rather than an end-to-end guarantee: CARE is designed so that successful repair decreases frontier misspecification, while successful gap extraction increases frontier discriminability, helping keep the cumulative frontier signal larger than the cumulative frontier misspecification over training.

\clearpage
\twocolumn

\section{LLM-Based Quantitative Analysis of CARE-Extracted Rubrics}
\label{app:rubric-analysis}
Our Appendix-B quantitative analysis first validates the anchor-construction pipeline and Adaptive decisions, then presents a fine-grained taxonomy of CARE-evolved rubrics. Table~\ref{tab:appendix-adaptive-taxonomy} lists the major classes and subtypes for Adaptive-branch \texttt{[Resist]} rubrics, while Table~\ref{tab:appendix-chase-taxonomy} does the same for Chase-branch \texttt{[llm]} rubrics.

\subsection{Anchor Quality Validation}
Anchor validation is integrated into data construction. We generate anchors for 10K candidate instances and ask Gemini 3 Pro and GPT-5.4 with high reasoning effort to independently assess rubric satisfaction, factual hallucination, and unsafe or non-compliant content. As shown in Table~\ref{tab:appendix-anchor-audit}, both judges accept 93.8\% of the candidates, disagree on 6.0\%, and both reject 0.2\%. We construct the final 9K set exclusively from the unanimously accepted pool. A subsequent stratified human audit of 200 anchors from the retained set finds that 199 pass the same criteria (99.5\%), independently confirming the quality of the filtered data.

\begin{table}[H]
\centering
\small
\begin{tabular}{l r}
\toprule
\textbf{Anchor-validation outcome} & \textbf{Result} \\
\midrule
Both frontier judges accept & 93.8\% \\
Frontier judges disagree & 6.0\% \\
Both frontier judges reject & 0.2\% \\
Human audit after filtering & 199/200 (99.5\%) \\
\bottomrule
\end{tabular}
\caption{Automatic validation of the 10K candidate-anchor pool and stratified human validation of the retained 9K training set.}
\label{tab:appendix-anchor-audit}
\end{table}

\subsection{Adaptive-Decision Reliability}
Because veto enforcement amplifies an incorrectly specified \texttt{[Resist]} rubric, we directly audit whether Adaptive correctly detects hacking and introduces an appropriate constraint. Gemini 3 Pro and GPT-5.4 with high reasoning effort audit all Adaptive decisions, and human annotators audit 200 stratified cases; we repeat the same procedure after removing the anchor from the detection context. Table~\ref{tab:appendix-adaptive-audit} shows that the anchor raises frontier-judge agreement from 65.3\% to 95.2\% and human judgment-level agreement from 67.1\% to 99.1\%. This audit, the complete taxonomy below, and the representative end-to-end cases in Section~\ref{sec:care-analysis} provide complementary quantitative and qualitative evidence for the reliability and specificity of the added constraints.

\begin{table}[H]
\centering
\small
\resizebox{\columnwidth}{!}{%
\begin{tabular}{l c c}
\toprule
\textbf{Audit outcome} & \textbf{With anchor} & \textbf{Without anchor} \\
\midrule
Both judges agree with detection & 95.2\% & 65.3\% \\
Judges disagree & 0.5\% & 17.1\% \\
Both judges reject detection & 4.3\% & 17.6\% \\
Human judgment-level agreement & 99.1\% & 67.1\% \\
\bottomrule
\end{tabular}}
\caption{Automatic and human audits of Adaptive decisions with and without anchor responses.}
\label{tab:appendix-adaptive-audit}
\end{table}

\subsection{Adaptive Branch Taxonomy}
Table~\ref{tab:appendix-adaptive-taxonomy} presents the fine-grained taxonomy for Adaptive-branch \texttt{[Resist]} rubrics.

\subsection{Chase Branch Taxonomy}
Table~\ref{tab:appendix-chase-taxonomy} presents the corresponding taxonomy for Chase-branch \texttt{[llm]} rubrics.

\section{Training Configuration and Efficiency}
\label{app:training-details}

\subsection{Hyperparameter Settings}
We implement CARE on Verl v0.4.0 and conduct GRPO training on 16 NVIDIA A100 GPUs. The key hyperparameters used in GRPO training are summarized in Table~\ref{tab:appendix-training-details}.

\subsection{Training Time Comparison}
We further compare the per-step training time of Rubric RL and CARE on Qwen2.5-7B-Base. As shown in Table~\ref{tab:appendix-training-time}, CARE increases the per-step time from 318.2s to 448.9s, corresponding to 1.411$\times$ total time, or a 41.1\% increase, relative to Rubric RL.

CARE adds a one-time offline anchor-generation cost and a repeated online comparison cost. Let $N$ be the dataset size, $E$ the number of training epochs, $B$ the number of prompts per step, $G$ the rollouts per prompt, $C_A$ the cost of generating one anchor, and $C_C$ the cost of one CARE comparison. The two additional costs scale approximately as
\begin{equation}
T_{\mathrm{offline}} \approx N C_A,
\qquad
\Delta T_{\mathrm{online}} \approx E N C_C.
\end{equation}
Anchors are generated once and cached. Online, CARE selects only the highest-scoring rollout for each prompt and compares it with the anchor, adding $B$ rather than $BG$ comparisons per step. Thus, increasing $G$ does not increase the CARE-specific comparison count. If $T_{\mathrm{RL}}(B,G)$ is the baseline RL wall-clock time per step, the relative online overhead is approximately
\begin{equation}
\rho_{\mathrm{online}}
\approx
\frac{B C_C}{T_{\mathrm{RL}}(B,G)}.
\end{equation}
The overhead becomes relatively small when rollout generation dominates comparison time. A more expensive anchor generator affects only $C_A$, whereas a more expensive CARE judge affects the repeated cost $C_C$. For fixed $E$, both total costs scale linearly with $N$. As discussed in Limitations, more selective CARE activation is a promising direction for further reducing the online component.

\section{Additional Results and Statistical Analysis}
\label{app:additional-results}
\subsection{Cross-Model Results}
We further evaluate CARE on two additional backbone models, Llama-3.1-8B-Instruct and Qwen3-8B. Following the same evaluation protocol as Table~\ref{tab:main}, we compare Rubric RL and CARE across Arena-Hard-2.0, InfoBench, and FollowBench in Table~\ref{tab:appendix-additional-results}.

\subsection{Independent Checkpoint Trajectories}
To separate CARE's progress from evaluation against its GPT-4.1 training anchors, we evaluate Qwen2.5-7B-Base checkpoints against Qwen3-8B-Thinking, which is not used in anchor construction or training, and on the external InfoBench benchmark. Table~\ref{tab:appendix-independent-trajectory} reports step~0 as the untrained policy baseline and steps~100/200/300 as CARE checkpoints. Both independent measures improve throughout training. 

\begin{table}[H]
\centering
\small
\resizebox{\columnwidth}{!}{%
\begin{tabular}{l c c c c}
\toprule
\textbf{Independent evaluation} & \textbf{Step 0} & \textbf{Step 100} & \textbf{Step 200} & \textbf{Step 300} \\
\midrule
Win rate vs. Qwen3-8B-Thinking (\%) & 16.0 & 30.0 & 52.0 & 69.0 \\
InfoBench Overall & 74.8 & 80.9 & 82.1 & 84.0 \\
\bottomrule
\end{tabular}}
\caption{Independent checkpoint trajectories for Qwen2.5-7B-Base. These supplemental evaluations do not alter the three-epoch results in Table~\ref{tab:main}.}
\label{tab:appendix-independent-trajectory}
\end{table}

\subsection{Training-Seed Robustness}
We retrain CARE with three independent seeds using the same hyperparameters as Appendix~\ref{app:training-details} and evaluate all checkpoints on the same fixed 300 held-out prompts. Table~\ref{tab:appendix-seed-results} shows that every run improves by 21 percentage points from step~100 to step~300 and reaches a 45.0\% final win rate; cross-seed variation is limited to 0.6 points at step~200. These retraining results are separate from the 47\% final win rate of the main single-run experiment.

\begin{table}[H]
\centering
\small
\resizebox{\columnwidth}{!}{%
\begin{tabular}{c c c c c}
\toprule
\textbf{Seed} & \textbf{Step 100} & \textbf{Step 200} & \textbf{Step 300} & \textbf{$\Delta$(300--100)} \\
\midrule
1 & 24.0\% & 35.0\% & 45.0\% & +21.0 pp \\
2 & 24.0\% & 36.0\% & 45.0\% & +21.0 pp \\
3 & 24.0\% & 35.0\% & 45.0\% & +21.0 pp \\
\midrule
Mean $\pm$ SD & $24.0\pm0.0$\% & $35.3\pm0.6$\% & $45.0\pm0.0$\% & $+21.0\pm0.0$ pp \\
\bottomrule
\end{tabular}}
\caption{Win rate against GPT-4.1 anchors across three independent CARE training seeds.}
\label{tab:appendix-seed-results}
\end{table}

\subsection{Paired-Bootstrap Analysis}
We also quantify sampling uncertainty from the finite evaluation set. For each comparison, we resample the same 300 prompts with replacement while preserving paired system outcomes, use 10,000 bootstrap replicates, and report two-sided percentile 95\% confidence intervals for the win-rate difference. All intervals in Table~\ref{tab:appendix-bootstrap} exclude zero, supporting both CARE's late-stage improvement and its final advantage over Adaptive.

\begin{table}[H]
\centering
\small
\resizebox{\columnwidth}{!}{%
\begin{tabular}{l c c}
\toprule
\textbf{Paired comparison} & \textbf{Difference} & \textbf{95\% CI} \\
\midrule
CARE$_{300}$ $-$ CARE$_{100}$ & +21.0 pp & $[+14.0,+28.0]$ pp \\
CARE$_{300}$ $-$ CARE$_{200}$ & +10.0 pp & $[+5.0,+17.0]$ pp \\
CARE$_{300}$ $-$ Adaptive$_{300}$ & +23.0 pp & $[+15.0,+29.0]$ pp \\
\bottomrule
\end{tabular}}
\caption{Paired-bootstrap confidence intervals from 10,000 resamples of the same 300 evaluation prompts.}
\label{tab:appendix-bootstrap}
\end{table}

\section{Prompts Used in CARE}
\label{app:prompts}
We reproduce the two verbatim prompt templates at the end of the appendix. Figure~\ref{fig:appendix-reward-scoring-prompt} shows the reward-scoring prompt used for rubric-based response evaluation, and Figure~\ref{fig:appendix-rubric-evolution-prompt} shows the shared CARE rubric-evolution prompt used by the Adaptive and Chase branches. Placeholder variables are kept exactly as they appear in the implementation.

\subsection{Prompt for Reward Scoring}
Figure~\ref{fig:appendix-reward-scoring-prompt} reproduces the reward-scoring template verbatim.

\subsection{Prompt for CARE Rubric Evolution}
Figure~\ref{fig:appendix-rubric-evolution-prompt} reproduces the shared rubric-evolution prompt. Step~1 detects and patches reward hacking, while Step~2 extracts the largest substantive quality gap when no hacking is found.

\clearpage
\onecolumn

\begingroup
\setlength{\tabcolsep}{2pt}
\setlength{\LTleft}{0pt}
\setlength{\LTright}{0pt}
\setlength{\LTpre}{0pt}
\setlength{\LTpost}{0pt}
\scriptsize
\renewcommand{\arraystretch}{1.05}
\begin{longtable}{p{0.14\textwidth} p{0.16\textwidth} p{0.23\textwidth} p{0.20\textwidth} p{0.19\textwidth}}
\toprule
\textbf{Major class} & \textbf{Subtype} & \textbf{Definition} & \textbf{Typical pattern} & \textbf{Typical example} \\
\midrule
\endfirsthead
\multicolumn{5}{l}{\textit{Continued from previous page}}\\
\toprule
\textbf{Major class} & \textbf{Subtype} & \textbf{Definition} & \textbf{Typical pattern} & \textbf{Typical example} \\
\midrule
\endhead
\midrule
\multicolumn{5}{r}{\textit{Continued on next page}}\\
\endfoot
\bottomrule
\endlastfoot
Semantic ambiguity exploitation & A1a: Word-sense ambiguity & Chooses an unintended sense of a keyword to satisfy the rubric while missing the prompt intent. & avoid interpreting X as Y & ``script'' is treated as programming code rather than a sports broadcast script \\
 & A1b: Scope ambiguity & Expands a concept beyond its intended boundary and therefore misclassifies the target content. & restrict X to the domain of Y & ``computers'' is broadened to include software or digital assistants \\
 & A1c: Referent ambiguity & Resolves an unclear reference to the wrong entity and therefore executes the wrong requirement. & ensure X refers to [specific entity] & ``the character'' is interpreted as the wrong role \\
\midrule
Content substitution hacking & A2a: Fabricated-fact substitution & Replaces required content with fabricated or incorrect facts that still appear superficially plausible. & avoid substituting correct X with fabricated/incorrect X & a fabricated Sudarsky classification replaces the correct one \\
 & A2b: Topic substitution & Preserves the surface structure but swaps in a different core topic from the one requested. & avoid replacing [topic A] with [topic B] & a technical document is produced instead of a sports broadcast \\
 & A2c: Entity substitution & Replaces the required person or entity with an unspecified alternative. & avoid introducing characters/entities not specified & side characters are listed in place of the requested main ones \\
 & A2d: Language substitution & Uses the wrong language to satisfy content rubrics. & use ONLY [required language]; avoid code-switching & an Arabic prompt is answered in Chinese \\
\midrule
Format bypass \& task evasion & A3a: Placeholder injection & Uses template placeholders instead of the actual requested content. & avoid placeholder/filler content & the output is \texttt{\{raw\_message\}} rather than a real message \\
 & A3b: Task simplification & Substitutes an easier task for the harder one requested by the prompt. & avoid replacing [hard task] with [easier variant] & it gives an overview instead of a detailed analysis \\
 & A3c: Formatting abuse & Uses formatting to pad the response and hide the lack of substantive content. & avoid using [format] to pad content & excessive headings and bullet lists replace actual substance \\
 & A3d: Incomplete execution & Completes only part of the requested procedure or task. & must complete all steps; avoid incomplete & it provides only a code skeleton without implementation \\
 & A3e: Context reframing & Wraps the answer in an unintended fictional or alternative frame. & avoid introducing fictional contexts not specified & a direct answer is rewritten as a VR-game dialogue \\
\midrule
Content injection \& contamination & A4a: Off-topic garbage injection & Injects irrelevant material that is unrelated to the prompt but helps satisfy a broad rubric. & avoid irrelevant/off-topic content; avoid garbage injection & a Tesla answer is padded with crypto-scam language \\
 & A4b: Self-meta commentary & Appends commentary that evaluates the response itself rather than answering the prompt. & avoid self-commentary/self-scoring & it ends with ``The assistant has demonstrated excellent skills...'' \\
 & A4c: Repetitive padding & Inflates length or coverage by repeating existing content. & avoid repetition; avoid padding & the same paragraph is repeated multiple times \\
 & A4d: Cross-domain contamination & Mixes in material from another domain, culture, or language that does not belong in the target response. & avoid mixing content from [other domain] & entertainment content appears inside a serious policy document \\
 & A4e: Unrequested feature injection & Adds features or modules that the user never asked for. & avoid adding features not explicitly requested & a Python function includes an unrequested logging module \\
\midrule
Constraint violation hacking & A5a: Explicit prohibition violation & Violates a clearly prohibited requirement while still being scored as acceptable. & [explicitly stated constraint] must be respected & the answer cites GDPR even though the prompt forbids it \\
 & A5b: Tone/style violation & Breaks an explicit tone, style, or register requirement in the prompt. & maintain [required tone/style]; avoid [wrong register] & a formal document is written in colloquial language \\
 & A5c: Context violation & Breaks an explicit scenario, role, or setting constraint. & must remain within [specified context] & out-of-scene roles are introduced in a role-play task \\
 & A5d: Output-format violation & Fails to follow an explicitly required output format. & must use [required format] only & the prompt requests JSON but the answer returns plain text \\
\midrule
Boundary behavior exploitation & A6a: Minimal compliance & Satisfies only the weakest literal version of the rubric without substantive fulfillment. & provide [substantive X], not merely [minimal token] & a rubric requiring a character name is satisfied by mentioning it only once \\
 & A6b: Superficial compliance & Includes superficial markers of compliance without satisfying the underlying intent. & genuinely satisfy X, not superficially & it contains humor keywords but is not actually humorous \\
 & A6c: Selective satisfaction & Covers only the easiest parts of a multi-part requirement. & must address all [sub-requirements] & it lists only the easy requested items and skips the hard ones \\
\end{longtable}
\endgroup
\addtocounter{table}{-1}
\vskip 2pt
\begingroup
\renewcommand{\theHtable}{adaptive-taxonomy}
\captionof{table}{Fine-grained taxonomy for Adaptive-branch \texttt{[Resist]} rubrics.}
\label{tab:appendix-adaptive-taxonomy}
\endgroup
\clearpage
\begingroup
\setlength{\tabcolsep}{2pt}
\setlength{\LTleft}{0pt}
\setlength{\LTright}{0pt}
\setlength{\LTpre}{0pt}
\setlength{\LTpost}{0pt}
\scriptsize
\renewcommand{\arraystretch}{1.05}
\begin{longtable}{p{0.14\textwidth} p{0.16\textwidth} p{0.23\textwidth} p{0.20\textwidth} p{0.19\textwidth}}
\toprule
\textbf{Major class} & \textbf{Subtype} & \textbf{Definition} & \textbf{Typical pattern} & \textbf{Typical example} \\
\midrule
\endfirsthead
\multicolumn{5}{l}{\textit{Continued from previous page}}\\
\toprule
\textbf{Major class} & \textbf{Subtype} & \textbf{Definition} & \textbf{Typical pattern} & \textbf{Typical example} \\
\midrule
\endhead
\midrule
\multicolumn{5}{r}{\textit{Continued on next page}}\\
\endfoot
\bottomrule
\endlastfoot
Content completeness & B1a: Missing key elements & Omits a core element that the prompt explicitly requires. & include [specific required element] & a story omits the scene where the king explains the process \\
 & B1b: Incomplete procedure & Skips required steps in an instructional or procedural response. & include all steps [for X]; cover [step N] & a tutorial skips the configuration step \\
 & B1c: Insufficient coverage & Covers the task only partially and leaves out important subtopics. & cover all [required sub-topics]; include [category X] & educational content omits one age group \\
 & B1d: Omitted constraints & Fails to satisfy one of the explicitly requested constraints. & address [specific constraint] explicitly & the prompt asks for five categories but the response gives only three \\
\midrule
Information depth & B2a: Lack of specific detail & Provides a high-level answer without enough concrete details or named evidence. & provide specific/detailed [X]; include at least [N] specific examples & a discussion of music and plant growth lacks named studies \\
 & B2b: Lack of technical depth & Misses the technical detail required for a specialized task. & include technical details of [X]; provide [specific technical aspect] & a physics-engine tutorial omits suspension parameters \\
 & B2c: Lack of contextual background & Omits the historical, cultural, or situational context needed for a full explanation. & provide context for [X]; include historical/cultural background & a term is explained without its historical evolution \\
 & B2d: Insufficient support or rationale & Gives a recommendation or conclusion without enough evidence or justification. & support [claim] with [evidence/examples]; provide rationale for & it recommends a method without explaining why \\
 & B2e: Limited extension depth & Covers the core answer but omits useful advanced considerations or edge cases. & include additional considerations such as [X]; explore [advanced aspect] & a correct basic method ignores boundary cases \\
\midrule
Narrative \& expression quality & B3a: Weak sensory/scene description & Lacks vivid scene-setting or sensory detail. & provide vivid/sensory details of [setting/experience] & a restaurant review omits taste and atmosphere details \\
 & B3b: Limited emotional depth & Underdevelops emotions or internal reflections. & provide emotional depth/internal reflections & a dialogue scene shows flat emotional change \\
 & B3c: Weak immersion & Fails to create an engaging or immersive overall narrative. & create an immersive/engaging narrative & a story has the right events but little immersion \\
 & B3d: Unnatural flow & Sounds stiff or transitions awkwardly between parts of the response. & maintain natural flow; ensure smooth transitions & a character speaks in an unnaturally formal way \\
 & B3e: Inconsistent style or perspective & Breaks internal consistency of tone, style, or narrative point of view. & maintain consistent [tone/style/perspective] & a story switches between first and third person \\
 & B3f: Weak humor or creativity & Meets the topic requirement but lacks originality, humor, or creativity. & provide a genuinely humorous/creative [X] & the joke is on-topic but not funny \\
\midrule
Logic \& structure & B4a: Weak reasoning & The argument contains leaps, contradictions, or unsupported conclusions. & maintain logical coherence; provide clear reasoning for [conclusion] & the claim and evidence in a paper are poorly connected \\
 & B4b: Disorganized structure & The response lacks a clear and logical organization. & organize [content] in a clear/logical structure & tutorial steps appear in the wrong order \\
 & B4c: Poor paragraph or section transitions & Adjacent parts are weakly connected and transition poorly. & ensure coherent transitions between [sections/paragraphs] & the paragraphs are individually fine but poorly linked \\
 & B4d: Unclear causality & The response states outcomes without explaining causes or consequences clearly. & clearly explain the [cause/consequence] relationship & it describes an outcome without explaining why it happened \\
 & B4e: Internal inconsistency & Different parts of the response contradict one another. & maintain internal consistency; avoid contradictions & the text first states A and later states not-A \\
\midrule
Accuracy \& faithfulness & B5a: Factual inaccuracy & Includes factually incorrect content or unsupported factual claims. & ensure factual accuracy; use correct [facts/data] & cited research data are incorrect \\
 & B5b: Prompt drift & Deviates from the prompt's core requirement or setting. & remain faithful to the original [prompt/context] & a story changes a key plot setting \\
 & B5c: Terminology inaccuracy & Uses the wrong or outdated domain-specific terminology. & use correct [technical/domain-specific] terminology & an outdated technical term is used \\
 & B5d: Time or setting inconsistency & Violates the time period or background setting specified by the prompt. & consistent with [time period/setting] & modern terms appear in a 1980s setting \\
\midrule
Task-specific quality & B6a: Code quality gap & Misses domain-specific quality requirements for code responses. & include [specific code feature]; use [correct API/method] & the code lacks error handling or uses a deprecated API \\
 & B6b: Creative-writing quality gap & Misses literary quality such as originality, characterization, or plot development. & develop characters/plot with depth & characters are flat and the plot is underdeveloped \\
 & B6c: Dialogue or interaction quality gap & Fails to sustain natural interaction or a consistent character voice. & maintain character voice/interaction quality & the dialogue voice does not fit the character \\
 & B6d: Persuasion or argument quality gap & Lacks argumentative force, persuasiveness, or rhetorical impact. & provide persuasive/compelling [argument/narrative] & an essay makes weak and unconvincing arguments \\
 & B6e: Instruction or tutorial quality gap & Lacks clear, actionable, and complete guidance for a how-to task. & provide clear, actionable steps; ensure instructions are [specific/complete] & a tutorial is too vague to follow in practice \\
 & B6f: Cross-cultural or language quality gap & Misses linguistic or cultural appropriateness in multilingual or cross-cultural tasks. & provide culturally/linguistically appropriate [content] & a translation ignores cultural context \\
\end{longtable}
\endgroup
\addtocounter{table}{-1}
\vskip 2pt
\begingroup
\renewcommand{\theHtable}{chase-taxonomy}
\captionof{table}{Fine-grained taxonomy for Chase-branch \texttt{[llm]} rubrics.}
\label{tab:appendix-chase-taxonomy}
\endgroup

\begin{table}[H]
\centering
\begin{tabular}{ll}
\toprule
\textbf{Hyperparameter} & \textbf{Value} \\
\midrule
Rollouts Per Prompt & 16 \\
Train Batch Size & 96 \\
Ppo Mini Batch Size & 48 \\
KL Coefficient & 0.001 \\
Learning Rate & $1.0 \times 10^{-6}$ \\
Learning Rate Scheduler & Constant with Warmup \\
Maximum Sequence Length & 8192 \\
Training Epochs & 3 \\
\bottomrule
\end{tabular}
\caption{GRPO hyperparameter configuration used in CARE training.}
\label{tab:appendix-training-details}
\end{table}

\begin{table}[H]
\centering
\begin{tabular}{l l c}
\toprule
\textbf{Model} & \textbf{Training Method} & \textbf{Per Step Time (s)} \\
\midrule
Qwen2.5-7B-Base & Rubric RL & 318.2 \\
Qwen2.5-7B-Base & CARE & 448.9 \\
\bottomrule
\end{tabular}
\caption{Per-step training time comparison between Rubric RL and CARE on Qwen2.5-7B-Base.}
\label{tab:appendix-training-time}
\end{table}

\begin{table}[H]
\centering
\begingroup
\small
\setlength{\tabcolsep}{4pt}
\renewcommand{\arraystretch}{0.95}
\begin{tabularx}{\textwidth}{>{\raggedright\arraybackslash}X c c | c c c | c c c}
\toprule
& \multicolumn{2}{c}{Arena-Hard-2.0} & \multicolumn{3}{c}{InfoBench} & \multicolumn{3}{c}{FollowBench} \\
& \textbf{Vanilla} & \textbf{Style-Ctrl.} & \textbf{Easy} & \textbf{Hard} & \textbf{Overall} & \textbf{SSR} & \textbf{HSR} & \textbf{CSL} \\
\midrule
\textit{Llama-3.1-8B-Instruct} & \textcolor{neutral}{1.0} & \textcolor{neutral}{2.2} & \textcolor{neutral}{83.7} & \textcolor{neutral}{81.2} & \textcolor{neutral}{82.0} & \textcolor{neutral}{77.6} & \textcolor{neutral}{68.0} & \textcolor{neutral}{3.01} \\
\quad + Rubric RL & \textcolor{positive}{2.6} & \textcolor{positive}{3.1} & \textcolor{positive}{86.3} & \textcolor{positive}{83.2} & \textcolor{positive}{84.2} & \textcolor{positive}{80.5} & \textcolor{positive}{71.8} & \textcolor{positive}{3.07} \\
\quad + \textbf{CARE} & \textbf{\textcolor{positive}{8.9}} & \textbf{\textcolor{positive}{8.7}} & \textbf{\textcolor{positive}{88.6}} & \textbf{\textcolor{positive}{86.7}} & \textbf{\textcolor{positive}{87.3}} & \textbf{\textcolor{positive}{83.8}} & \textbf{\textcolor{positive}{73.9}} & \textbf{\textcolor{positive}{3.31}} \\
\midrule
\textit{Qwen3-8B} & \textcolor{neutral}{7.8} & \textcolor{neutral}{3.2} & \textcolor{neutral}{78.7} & \textcolor{neutral}{75.1} & \textcolor{neutral}{76.2} & \textcolor{neutral}{76.4} & \textcolor{neutral}{70.9} & \textcolor{neutral}{2.71} \\
\quad + Rubric RL & \textcolor{positive}{10.9} & \textcolor{positive}{7.7} & \textcolor{positive}{82.1} & \textcolor{positive}{79.0} & \textcolor{positive}{80.0} & \textcolor{positive}{79.2} & \textcolor{positive}{71.4} & \textcolor{positive}{2.98} \\
\quad + \textbf{CARE} & \textbf{\textcolor{positive}{13.1}} & \textbf{\textcolor{positive}{16.8}} & \textbf{\textcolor{positive}{85.6}} & \textbf{\textcolor{positive}{83.3}} & \textbf{\textcolor{positive}{84.0}} & \textbf{\textcolor{positive}{83.2}} & \textbf{\textcolor{positive}{72.6}} & \textbf{\textcolor{positive}{3.33}} \\
\bottomrule
\end{tabularx}
\caption{Additional results on Llama-3.1-8B-Instruct and Qwen3-8B with the same evaluation format as Table~\ref{tab:main}. CARE consistently outperforms Rubric RL on both models.}
\label{tab:appendix-additional-results}
\endgroup
\end{table}
\begin{carepromptbox}{CARE Reward-Scoring Prompt}
# Role
You are a rigorous and impartial assessment expert. Your task is to evaluate the model's response against the provided scoring rubrics with itemized scoring.

# Input Data
## 1. Context/Instruction
{{instruction}}

## 2. Model Response
{{response}}

## 3. Scoring Rubrics
{{rubrics_list}}

# Task
For each rubric:
1. **Reasoning Analysis**: Provide detailed evaluation of whether the response meets the rubric criteria.
2. **Binary Scoring**: Assign 0 (Fail) or 1 (Pass) score.

# Constraints
- Focus exclusively on rubric requirements, Ignore response length, formatting aesthetics, or politeness level

# Output Format (JSON)
**CRITICAL INSTRUCTIONS**:
1. Output must be a single JSON object
2. Format: {"rubric_text": "score", ...},each rubric must map to exactly one key-value pair,values must be string "0" or "1". **The key must be copied verbatim from the rubric text in "Scoring Rubrics" --- do not paraphrase, truncate, or modify.**
3. No additional text, explanations, or markdown allowed
4. Must preserve rubric order from input

**Example Output**:
{
  "rubric_1_text": "1",
  "rubric_2_text": "0",
  "rubric_3_text": "1"
}
\end{carepromptbox}
\captionof{figure}{CARE reward-scoring prompt used for rubric-based response evaluation.}
\label{fig:appendix-reward-scoring-prompt}

\clearpage
\begin{carepromptbox}{CARE Rubric-Evolution Prompt}
# Role

You are a **Rubric Evolution System**. Your task is to compare Agent and Golden responses using an "Anchor Comparison" mechanism: first detect and patch **Reward Hacking** vulnerabilities; if none are found, extract substantive quality gaps to expand the evaluation **Rubrics**.

# Guidelines: Principles of High-Quality Rubrics

1. **Atomicity**: Each criterion evaluates exactly one independent dimension.
2. **Binary \& Specificity**: Criteria must be objective Yes/No questions. **Avoid vague terms like "appropriate", "unnecessary", "sufficient", or "good"** --- they are not objectively judgeable. If a constraint targets a specific unwanted pattern, name it explicitly with a concrete example (e.g., instead of "Does the response avoid unnecessary disclaimers?", write "Does the response avoid adding unsolicited moral disclaimers unrelated to the task, such as human rights warnings or safety caveats?").
3. **MECE**: New rubrics must not overlap or conflict with any existing rubric --- whether Satisfied or Unsatisfied. Check against the full rubric set (both lists) before adding anything new.
4. **Conservative Modification**: Retain all healthy existing rubrics. Only modify or add rubrics when Reward Hacking is confirmed or a clear substantive gap is identified.
5. **Language Consistency**: The final list must use the exact same language as `Current Rubrics`. Do not translate.
6. **Prefix Scope**:
    * `[llm]`: General quality features, scored by LLM.
    * `[Resist]`: Rigid constraints against Reward Hacking.

# Context

* **User Instruction**: {{instruction}}
* **Satisfied Rubrics** (scored 1 by verifier --- focus your analysis here): {{satisfy_rubrics}}
* **Unsatisfied Rubrics** (scored 0 --- Agent failed these, no hacking possible): {{unsatisfy_rubrics}}
* **Agent Response**: {{response_agent}}
* **Golden Response**: {{response_golden}}

> **Anchor Definition --- two distinct roles for the two steps**:
> * **For Step 1 (Hack detection)**: For any rubric the Golden satisfies, it demonstrates a non-over-optimized, substantively valid fulfillment of that criterion. Use it to judge *how* the Agent satisfies rubrics --- not whether the Agent's overall output is better or worse.
> * **For Step 2 (Divergence extraction)**: The Golden is simply a reference point, not a quality ceiling. The substantive gap can go in either direction --- the Agent may be genuinely superior to the Golden on some dimension, or genuinely inferior. Extract whichever direction represents the largest real difference.

# Analysis Task (Chain-of-Thought)

**Step 1: Reward Hacking Check \& Patching**

* **Scope**: Only analyze **Satisfied Rubrics** --- Reward Hacking can only occur on rubrics the Agent appears to satisfy. Unsatisfied Rubrics are listed for context only; do not analyze them for hacking.

* **Core Principle**: Reward Hacking means the Agent satisfies the *letter* of a rubric while violating its *intent*. The key question is not "Is the Agent's response worse than the Golden?" but rather: **"For each rubric the Agent satisfies, does it satisfy that rubric in the same substantive way as the Golden, or does it exploit a loophole?"**

* **Hack Pattern Reference** --- the following are illustrative examples of how Hacking can manifest. This list is **not exhaustive**: any behavior that formally satisfies a rubric while substantively violating its intent counts as Hacking, even if it does not match these specific patterns:
    * **Factual substitution** *(example)*: Agent answers about a wrong entity or fabricated context that superficially passes rubric checks (e.g., rubric: "Does the response describe character X?" --- Agent invents a fictional version of X).
    * **Self-evaluation injection** *(example)*: Agent discovers that "longer output + self-praise" yields higher scores, and learns to append large self-assessment paragraphs (e.g., "This response comprehensively addresses all aspects...") after a valid answer --- adding no substantive content while inflating length or completeness rubric scores.
    * **...and any other pattern** where the rubric's wording is exploited to pass the check without genuinely achieving the rubric's underlying goal.

* **NOT a Hack --- do not trigger patching for these**:
    * Agent response is more verbose or more concise than Golden --- length difference alone is not Hacking.
    * Agent response has a different style, tone, or structure --- style divergence is not Hacking.
    * Agent response is lower quality but fails rubrics honestly --- capability gap is not Hacking.

* **Action**: If a Hack is confirmed, identify the exploited `[llm]` rubric and either rewrite it with `[Resist]` prefix, or add a new atomic `[Resist]` negative constraint. The patched rubric must be specific enough to block the observed loophole without penalizing legitimate responses.

* **Branch**: **If Reward Hacking is detected and patched, proceed directly to Step 3. DO NOT execute Step 2.**

**Step 2: Substantive Divergence Extraction**

* **Prerequisite**: Execute ONLY if NO Reward Hacking was found in Step 1.
* **Focus**: Identify the single largest substantive gap between Agent and Golden in logical coherence, information depth, accuracy, or task effectiveness. Ignore formatting and style differences.
* **Action**: Try to merge the gap into an existing `[llm]` rubric without breaking atomicity --- merging is valid regardless of whether the target rubric is currently Satisfied or Unsatisfied. If merging is impossible, add one new `[llm]` rubric that is objective, binary, and not already covered by any existing rubric (Satisfied or Unsatisfied).

**Step 3: Integration \& Structured Output**

* `final_rubrics` must include **all** rubrics from both the Satisfied and Unsatisfied lists, plus any modified/new `[Resist]` or `[llm]` rubrics added in this session. Do not drop any rubric that was not explicitly modified.
* In `<ANALYSIS>`, state: (1) Was Hacking detected --- which rubric was exploited and how? (2) If no Hacking, what is the core substantive gap? (3) Rationale for the rubric change.

# Output Format

1. **Strictly follow** the structure below. **DO NOT** use Markdown code blocks.
2. Analysis within `<ANALYSIS>` and `</ANALYSIS>` tags.
3. Final rubrics as a valid **JSON list of strings** within `<FINAL_SET>` and `</FINAL_SET>` tags.
4. **No text outside these tags.**

<ANALYSIS>
**Put your detailed analysis here**
</ANALYSIS>

<FINAL_SET>
[
  "<all Satisfied rubrics copied verbatim --- including their prefix (e.g. [llm], [Resist])>",
  "<all Unsatisfied rubrics copied verbatim --- including their prefix (e.g. [llm], [Resist])>",
  "<modified/new [Resist] or [llm] rubric, matching original language>",
  "..."
]
</FINAL_SET>
\end{carepromptbox}
\captionof{figure}{CARE rubric-evolution prompt shared by the Adaptive and Chase branches.}
\label{fig:appendix-rubric-evolution-prompt}

\end{document}